\documentclass{IEEEtran}
\pdfoutput=1

\PassOptionsToPackage{numbers,sort&compress}{natbib}

\usepackage[utf8]{inputenc} % allow utf-8 input
\usepackage[T1]{fontenc}    % use 8-bit T1 fonts
\usepackage[draft]{hyperref}       % hyperlinks
\usepackage{url}            % simple URL typesetting
\usepackage{booktabs}       % professional-quality tables
\usepackage{amsfonts}       % blackboard math symbols
\usepackage{nicefrac}       % compact symbols for 1/2, etc.
\usepackage{microtype}      % microtypography

\usepackage{bm}
\usepackage{url}
\usepackage{tikz}
\usepackage{float}
\usepackage{amsmath}
\usepackage{amssymb}
\usepackage{amsthm}
\usepackage{authblk}
\usepackage{xspace}
\usepackage{color}
\usepackage{graphicx}
\usepackage{grffile}
\usepackage{soul}
\usepackage{etoolbox}
\usepackage{chngpage}
\usepackage{environ}
\usepackage{subcaption}
\usepackage{rotating}
\usepackage{multirow}
\usepackage{algorithm}
\usepackage{algorithmicx}
\usepackage[noend]{algpseudocode}

\usepackage[numbers]{natbib}

\newcommand{\newterm}[1]{{\it #1}}

\def\Figref#1{Figure~\ref{#1}}

\def\Secref#1{Section~\ref{#1}}
\def\eqref#1{equation~\ref{#1}}
\def\Eqref#1{Equation~\ref{#1}}
\def\1{\bm{1}}

\def\va{{\bm{a}}}
\def\vb{{\bm{b}}}

\def\vv{{\bm{v}}}
\def\vw{{\bm{w}}}
\def\vx{{\bm{x}}}

\def\vz{{\bm{z}}}

\DeclareMathAlphabet{\mathsfit}{\encodingdefault}{\sfdefault}{m}{sl}
\SetMathAlphabet{\mathsfit}{bold}{\encodingdefault}{\sfdefault}{bx}{n}

\def\gG{{\mathcal{G}}}

\def\sR{{\mathbb{R}}}
\def\sS{{\mathbb{S}}}

\def\sX{{\mathbb{X}}}

\newtoggle{shortpaper}

\newcommand{\descr}[1]{\paragraph{#1}}

\newcommand{\astar}{A$^*$\xspace}
\newcommand{\nope}{$\times$}
\newcommand{\yup}{$\checkmark$}

\def\Stmtref#1{Statement~\ref{#1}}

\def\Appref#1{Appendix~\ref{#1}}

\def\Tabref#1{Table~\ref{#1}}

\newcommand{\lp}[1][p]{\mathcal{L}_#1}
\newcommand{\conflevel}{l}
\newcommand{\probthresh}{d}
\newcommand{\scorethresh}{\theta}

\newcommand{\edgecost}{\omega}

\newcommand{\cost}{C}
\newcommand{\pathcost}{W}
\newcommand{\graphcost}{C_{\gG}}
\newcommand{\domainclosure}{\sS}
\newcommand{\reprspace}{\sR^m}
\newcommand{\optim}[1]{#1^{*}}

\newcommand{\hinputspace}{h}
\newcommand{\dist}[1][]{r_{#1}}
\newcommand{\distlowbound}[1][]{\eta_{#1}}
\newcommand{\openset}{\textsc{open}\xspace}
\newcommand{\closedset}{\textsc{closed}\xspace}
\newcommand{\isgoal}{\mathsf{goal}\xspace}
\newcommand{\scorefunc}{\mathsf{score}\xspace}
\newcommand{\opensetstruct}{\mathsf{pqueue}\xspace}
\newcommand{\norm}[1]{\left\lVert#1\right\rVert}
\newcommand{\dualnorm}[1]{\norm{#1}^*}

\newcommand*\Let[2]{\State #1 $\gets$ #2}

\title{Evading Classifiers in Discrete Domains with Provable Optimality Guarantees}

\author[1]{Bogdan Kulynych}
\author[2]{Jamie Hayes}
\author[1,3]{Nikita Samarin}
\author[1]{Carmela Troncoso}

\affil[1]{EPFL SPRING Lab}
\affil[2]{University College London}
\affil[3]{University of California, Berkeley}

\newtheorem{statement}{Statement}[section]
\newtheorem{definition}{Definition}[section]

\theoremstyle{remark}
\newtheorem{example}{Example}

\togglefalse{shortpaper}

\begin{document}

\maketitle

% !TEX root = ../oakland/main.tex
\begin{abstract}
    Machine-learning models for security-critical applications such as bot, malware, or spam
    detection, operate in constrained discrete domains. These applications would benefit from having
    provable guarantees against adversarial examples. The existing literature on provable
    adversarial robustness of models, however, exclusively focuses on robustness to gradient-based
    attacks in domains such as images. These attacks model the adversarial cost, e.g., amount of
    distortion applied to an image, as a $p$-norm. We argue that this approach is not well-suited to model
    adversarial costs in constrained domains where not all examples are feasible.

    We introduce a graphical framework that (1) generalizes existing attacks in discrete domains,
    (2) can accommodate complex cost functions beyond $p$-norms, including financial cost
    incurred when attacking a classifier, and (3) efficiently produces valid adversarial examples with
    guarantees of minimal adversarial cost. These guarantees directly translate into a notion of
    adversarial robustness that takes into account domain constraints and the adversary's
    capabilities. We show how our framework can be used to evaluate security by crafting
    adversarial examples that evade a Twitter-bot detection classifier with provably minimal
    number of changes; and to build privacy defenses by crafting adversarial examples that evade a
    privacy-invasive website-fingerprinting classifier.
\end{abstract}

% !TEX root = ../oakland/main.tex

\section{Introduction}
Many classes of machine-learning (ML) models are vulnerable to efficient gradient-based attacks that
cause classification errors at test time~\cite{MadryMSTV17, CarliniWagner17, Moosavi-Dezfooli16,
PapernotMJFCS16, GoodfellowSS14, BiggioCMNSLGR13, SzegedyZSBEGF13}. A large body of work has been
dedicated to obtaining \newterm{provable guarantees of adversarial robustness} of ML classifiers
against such attacks~\cite{BastaniILVNC16, KatzBDJK17, HeinA17, FawziFF18, WongSMK18, TsuzukuSS18}.
These works focus exclusively on continuous domains such as images, where an attacker adds small
\newterm{perturbations} to regular examples such that the resulting \newterm{adversarial examples}
cause a misclassification. Their definition of robustness is that the classifier's decision is
stable in a certain $\lp$-neighbourhood around a given example, i.e., perturbations having an $\lp$
norm lower than a threshold cannot flip the decision of the classifier.

Security-critical applications of machine learning such as bot, malware, or spam detection rely on
feature vectors whose values are constrained by the specifics of the problem. In these settings,
adding small perturbations to examples could result in feature vectors that cannot appear in real
life~\cite{DemontisMBMARCG17, EbrahimiRLD18, JiaG18, KolosnjajiDBMGER18}. For example, perturbing a
malware binary to prevent antivirus detection could turn the binary non-executable; or perturbing a
text representation to change the output of a classifier~\cite{MiyatoDG16} could cause the
representation to not correspond to any plausible text in terms of semantics or grammar.
Hence, the techniques for evaluating robustness against perturbation-based adversaries 
are not straightforward to apply to these cases.

This shows a significant gap in the literature. On one hand, multiple methods and tools have been
proposed to either evaluate such measures of robustness, or train robust models. On the other hand,
many security-critical domains---that could benefit from such tools---cannot effectively use them,
as the measures do not easily translate to discrete domains.

% Moreover, if one were to find an adversarial example that takes into account non-trivial domain
% constraints, it could be too costly to produce in reality. For example, a bot operator may evade a
% Twitter bot detector by increasing the number of replies to their tweets.  Yet, this may require a
% large investment to either create or buy such replies. Thus, when crafting adversarial examples for
% security applications it is important to take into account the costs. The typical cost model used in
% existing methods, the $\lp$ norm of the perturbation, however, might not be suitable to capture the
% complex manipulation costs required in discrete domains~\cite{SharifBBR16, GilmerAGAD18}.

We introduce a framework for efficiently finding adversarial examples that is suitable for
constrained discrete domains. We represent the space of possible adversarial manipulations as a
weighted directed graph, called a \newterm{transformation graph}. Starting from a regular example,
every edge is a transformation, its weight being the transformation cost, and the children nodes are
transformed examples.

This representation has the following advantages. First, explicitly defining the descendants for
each node captures the feasibility constraints of the domain: the transitive closure for a given
starting node represents the set of all possible transformations of that example. Second, the graph
can capture non-trivial manipulation-cost functions: the cost of a sequence of manipulations can be
modeled as the sum of edge weights along a path from the original to the transformed example. Third,
the graph representation is independent of the ML model being attacked, and of the adversary's
knowledge of this model. Thus, it applies to different adversarial settings.  Fourth, this framework is
a generalization of many existing attacks in discrete domains~\cite{PapernotMJFCS16, PapernotMSH16,
GrossePM0M16, EbrahimiRLD18, LiangSBLS18, GaoLSQ18, JiaG18, OverdorfKBTG18}. This makes it a
useful tool for comparing attacks and characterizing the attack space. 

An additional advantage of the graphical approach is that it enables us to use well-known algorithms
for graph search in order to find adversarial examples. Concretely, in a white-box setting, an
adversary can use the \astar graph-search algorithm to find adversarial examples that are
\emph{optimal} in terms of transformation cost, i.e., with \newterm{minimal adversarial cost}
guarantees. Note that we use the term \emph{adversarial cost} to represent the effort an adversary
applies to mount an attack~\cite{LowdM05}. Works on \newterm{cost-sensitive adversarial
robustness}~\cite{AsifXBZ15, ZhangE19} use the term \emph{cost} in a different sense---to represent
the harm an adversary causes with different kinds of misclassifications in a multi-class
setting---and, hence, are orthogonal to our work.

Being able to obtain constrained adversarial examples with provably minimal costs has key
implications for security. The minimal adversarial cost guarantee naturally extends the notion of
adversarial robustness in an $\lp$-neighbourhood to constrained discrete domains. Furthermore, our
approach enables the evaluation of adversarial robustness under realistic tangible cost functions.
For instance, we show how to measure robustness in terms of economical cost, i.e., guaranteeing that
the adversary needs to pay a certain financial price in order to change the decision of the
classifier.

Using \astar search to find minimal-cost adversarial examples requires to compute a heuristic based
on a measure of adversarial robustness in a $\lp$-neighbourhood over a continuous superset of the
domain. Hence, we establish the following connection: if adversarial robustness of a classifier can
be computed in a $\lp$-neighborhood in a continuous domain, it can also be computed in a discrete
domain using our framework. If computing robustness in the continuous domain is expensive, we show
how efficient sub-optimal instantiations of \astar can be used to still obtain optimality guarantees
on the costs.

The graphical framework and the provable guarantees it enables to obtain are also useful in privacy
applications. Machine learning is widely used to infer private information about
users~\cite{Cadwalladr18, AbbasiC08, KosinskiSG13}, track people~\cite{HRW18}, and learn about their
browsing behaviour~\cite{BackMS01,LiberatoreL06}. Defenses against such attacks are non-existent or
predominantly ad-hoc (e.g., for de-anonymization attacks \cite{PanchenkoNZE11, DyerCRS12, CaiNWJG14,
CaiNJ14, JuarezIPDW17}). With the exception of the work by \citeauthor{JiaG18} on hindering
profiling based on app usage~\cite{JiaG18}, the privacy community has so far not considered the use of 
evasion attacks as systematic means to counteract privacy-invasive classifiers. As
many of the domains in which privacy-invasive classifiers operate are discrete, our framework
opens the door to the principled design of privacy defenses against such classifiers. Moreover, the
minimal cost guarantee provides a lower bound on the costs of a defense. Thus, it provides a good
baseline for benchmarking the cost-effectiveness of existing privacy defenses against machine
learning.

\smallskip

In summary, these are our contributions:
\begin{itemize}
    \item We present a graphical framework that systematizes the crafting of adversarial examples in
        constrained discrete domains. It generalizes many existing crafting methods.
    \item We show how to use the framework to measure adversarial robustness considering the domain
        constraints and the adversary's capabilities. Our framework can handle arbitrary
        costs beyond the commonly-used $\lp$ norms to express the adversarial costs.
    \item We show how the \astar graph search algorithm can be used to efficiently obtain minimal-cost
        adversarial examples, thus to provide provable guarantees of robustness against a given model of
        adversarial capabilities.
    \item We identify and formally prove the connection between adversarial robustness in continuous
        and discrete domains: if the robustness can be computed in the continuous domain, 
        it can also be computed in a discrete domain.
    \item We show how our framework can be used as a tool to systematically build and evaluate privacy
        defenses against privacy-invasive ML classifiers.
\end{itemize}

% !TEX root = ../oakland/main.tex

\section{A Graph Search Approach to Evasion}\label{sec:model}
After some preliminaries, we introduce our graphical framework for designing evasion attacks.

\subsection{Preliminaries}\label{sec:model:prelims}
Throughout the paper we denote vectors in $\sR^m$ using bold face: $\vx$. We denote by $\va
\cdot \vb$ a dot product between two vectors. 

\subsubsection{Binary Classifiers}
In this work, we focus on binary \newterm{classifiers}, $F: \sX \rightarrow \{0, 1\}$ that produce a
decision $\{0, 1\}$ by thresholding a \newterm{discriminant function} $f(x)$:
    \[
        F(x) = \begin{cases} 1, & f(x) > \scorethresh \\ 0, & \text{otherwise} \end{cases}
    \]
where $\scorethresh \in \sR$ is a decision threshold. The discriminant function $f(x) = \vw \cdot
\phi(x) + b$ is a composition of a possibly non-linear \newterm{feature mapping} $\phi: \sX
\rightarrow \reprspace$ from some input space $\sX$ to a feature space $\reprspace$, and a linear
function $\vw \cdot \vz + b$ with $\vz = \phi(x)$. This encompasses several families of models 
in machine learning, including logistic regression, SVM, and neural network-based classifiers.
In the rest of this paper we use the terms \emph{classifier} and \emph{model} interchangeably.

Often, % (e.g., in logistic regression)
the decision threshold is defined through a \newterm{confidence} value $\probthresh \in [0, 1]$ such
that $\probthresh = \sigma(\scorethresh)$, that is, $\scorethresh = \sigma^{-1}(\probthresh)$, where
$\sigma: \sR \rightarrow [0, 1]$ is a sigmoid function: \[\sigma(y) = \frac{1}{1 + e^{-y}}\]

Binary classifiers are often employed in security settings for detecting security violations.
Some standard examples are spam, fraud, bot, or network-intrusion detection.
%, where the classifier outputs ``no violation'' or ``violation detected.''

\subsubsection{Graph Search}
\label{sec:bfs}
Let $\gG = (V, E, \edgecost)$ be a directed weighted graph, where $V$ is a set of nodes, $E$ is a
set of edges, and $\edgecost: E \rightarrow \sR^+$ associates each edge with a weight, or
\newterm{edge cost}. For a given path $x_1 \rightarrow x_2 \rightarrow \ldots \rightarrow x_n$ in
the graph, we define the \newterm{path cost} as the sum of its edges' costs: \[\pathcost(x_1
\rightarrow x_2 \rightarrow \ldots \rightarrow x_n) = \sum_{i=1}^{n-1} \edgecost(x_i, x_{i + 1}).\]

Let $\isgoal: V \rightarrow \{\top, \bot\}$ be a \newterm{goal predicate}. For a given starting node
$s \in V$ a \newterm{graph search algorithm} aims to find a node $g \in V$ that satisfies
$\isgoal(g) = \top$ such that the cost of reaching $g$ from $s$ is minimal:
\[ g = \arg \min_{v \in V} \graphcost(s, v) \text{ s.t. } \isgoal(v) = \top, \]
where $\graphcost(x, x')$ is defined as the minimal path cost over all paths in graph $\gG$ from $s$ to $v$:
\begin{equation}\label{eq:graph-cost}
    \begin{aligned}
        & \graphcost(s, v) = \min\limits_{v_i \in V} \pathcost(s \rightarrow v_1 \rightarrow
        \ldots \rightarrow v_{n-1} \rightarrow v) \\
        \text{ s.t. } & (s, v_1), (v_{n-1}, v), (v_i, v_{i+1}) \in E \quad \forall{i \in
        \overline{1, n-1}}
    \end{aligned}
\end{equation}

We call a global minimizer $g$ an \newterm{optimal}, or \newterm{admissible}, solution to the graph
search problem.

In Algorithm~\ref{alg:bfs}, we show the pseudocode of the BF$^*$ graph-search
algorithm~\cite{DechterP85}. Some common graph-search algorithms are specializations of BF$^*$:
uniform-cost search (UCS)~\cite{HartNR68}, greedy best-first search~\cite{DoranMichie66}, \astar and
some of its variants~\cite{HartNR68, Pohl70}. They differ in their instantiation of the scoring
function used to select the best nodes at each step of the algorithm. Additionally, by limiting the
number of items in the data structure holding candidate nodes, beam-search and hill-climbing
variations of the above algorithms can be obtained~\cite{RichKnight91}. We summarize these
differences in~\Tabref{tab:search-algos}.

\begin{algorithm}[t]
\caption{BF$^*$ search algorithm}\label{alg:bfs}
  \begin{algorithmic}[1]
	\Require{Priority queue data structure $\opensetstruct$}
	\Require{Directed graph $\gG = (V, E, \edgecost)$}
	\Require{Scoring function $\scorefunc: V \times V \rightarrow \sR$}
	\Require{Goal predicate $\isgoal: V \rightarrow \{0, 1\}$}
	\Require{Starting node $s \in V$}
    \Function{BF$^*$}{$\gG$, $\scorefunc(\cdot)$, $\isgoal(\cdot)$; $s$}
        \Let{\openset}{$\opensetstruct(\{s\})$}
		\Let{\closedset}{$\{\}$}
		\While{\openset is not empty}
			\Let{$v$}{remove node with lowest $f$-score from \openset}
			\If{$\isgoal(v)$}
				\Return $v$
			\EndIf
			\Let{\closedset}{$\closedset \cup \{v\}$}
			\For{each child $v'$ of $v$ in $\gG$}
				\Let{$\scorefunc$-value}{$\scorefunc(v, v')$}
				\If{$v'$ \emph{not} in \openset or \closedset}
					\State{Record $v'$ in \openset with $\scorefunc$-value}
				\EndIf
				\If{$v'$ is in \openset or \closedset and $\scorefunc$-value
                    \phantom{\hspace{5.2em}} is lower than recorded}
					\State{Replace $v'$ with the updated}
                    \State{\quad $\scorefunc$-value in the respective set}
					\If{$v'$ is in \closedset}
						\State{Move $v'$ to \openset}
					\EndIf
				\EndIf
			\EndFor
		\EndWhile
    \EndFunction
  \end{algorithmic}
\end{algorithm}

\begin{table}[t]
    \centering
    \caption{Specializations of BF$^*$. $h: V \rightarrow \sR$ is a heuristic function that
    estimates the path cost to reach a goal node.}\label{tab:search-algos}
    \resizebox{\columnwidth}{!}{
    \begin{tabular}{ll}
        \toprule
        \textbf{Algorithm} & \textbf{$\scorefunc$} \\
        \midrule
        Greedy best-first~\cite{DoranMichie66}
        & $h(v')$
        \\

        Uniform-cost~\cite{HartNR68}
        & $\edgecost(v, v')$
        \\

        \astar~\cite{HartNR68}
        & $\edgecost(v, v') + h(v')$
        \\

        $\varepsilon$-weighted \astar~\cite{Pohl70}
        & $\edgecost(v, v') + \varepsilon h(v')$
        \\

        \bottomrule

        \\

        \toprule
        \textbf{Algorithm} & \textbf{$\opensetstruct$} \\
        \midrule
        Hill climbing
        & Limited to one best-scoring item
        \\

        Beam search~\cite{RichKnight91}
        & Limited to $B$ best-scoring items
        \\
        \bottomrule
    \end{tabular}
    }
\end{table}

\subsection{The Graphical Framework}

\subsubsection{The Adversary's Strategy and Goal}
\label{sec:model:goal}
We assume the adversary relies on the ``mimicry'' strategy~\cite{DemontisMBMARCG17} to
\newterm{evade} an ML classifier: Departing from a known initial example $x$, the adversary applies
structure-preserving \newterm{transformations} until a transformed \newterm{adversarial example},
$x'$, is misclassified.

The adversary also wants to minimize the cost of these transformations. This problem is often
formulated as an optimization problem:
\begin{equation}\label{eq:mincost-problem}
    \optim{x} = \arg \min_{x' \in \sX} \cost(x, x') \text{ s.t. } \isgoal(x') = \top,
\end{equation}
where $x$ is the initial example, and $\cost(x, x') > 0$ is the \newterm{adversarial cost}.  $\cost$
models the ``price'' that the adversary pays to transform example $x$ into~$x'$. The adversary's goal
in this problem is to cause a misclassification error with a certain confidence level $\conflevel
\geq \probthresh$:
\begin{equation}\label{eq:target-conf-goal}
    \isgoal(x') = \begin{cases}
        \top, & t = 1 \text{ and } \sigma(f(x')) > \conflevel \\
        \top, & t = 0 \text{ and } \sigma(f(x')) \leq 1-\conflevel \\
        \bot, & \text{otherwise}
    \end{cases}
\end{equation}
where $t$ is the target class which is different from the original class $F(x)$ (if $F(x)=0$, then
$t=1$, and vice-versa).

If $\conflevel$ is equal to the decision threshold of the classifier, $\conflevel = \probthresh$, the
adversary merely aims to flip the decision. The adversary might also want not only to flip the
decision, but to make adversarial examples that are classified with higher confidence. This
corresponds to higher confidence levels: $\conflevel > \probthresh$.

\subsubsection{The Adversary's Knowledge}
Following standard practices for evaluating security properties of ML models, we assume the
worst-case, \newterm{white-box}, adversary that has full knowledge of the target model parameters,
including $\vw, b$ and the feature mapping~$\phi$. In \Secref{sec:eval-wfp} we also discuss attacks that
are applicable to a \newterm{black-box setting}, where the adversary does not have knowledge of the
model parameters or architecture, but can query it with arbitrary examples $x$ to obtain $f(x)$.

\subsubsection{The Adversary's Capabilities}

We model the capabilities of the adversary, including inherent domain constraints
and the cost of modifications, using a \newterm{transformation graph} that encodes 
the transformations the adversary can perform on an example $x$. 
This graph has to be defined \emph{before} running the attack.

The transformation graph is a directed weighted graph $\gG = (V, E, \edgecost)$, with $V \subseteq
\sX$ being a subset of the model's input space that the adversary can craft. An edge $(x, x') \in E$
represents the transformation of an example $x$ into an example $x'$. For each edge $(x, x') \in E $
the function $\edgecost$ defines the cost $\edgecost(x, x') > 0$ associated with that
transformation. A path cost $C(x_1 \rightarrow x_2 \rightarrow \ldots \rightarrow x_n)$ represents
the cost of performing a chain of transformations $x_1 \rightarrow x_2 \rightarrow \ldots
\rightarrow x_n$.

\subsubsection{Graphical Formulation}
Within the graphical framework, the problem in \Eqref{eq:mincost-problem} is reduced to minimizing
the transformation cost as defined by the graph $\gG$, thus narrowing the search space to only those $x'$
that are reachable from $x$:
\begin{equation}\label{eq:mincost-graph-problem}
    \begin{aligned}
        & \optim{x} = \arg \min_{x' \in V} \graphcost(x, x') \\
        \text{ s.t. } & \isgoal(x') = \top \\
        & \text{$x'$ is reachable from $x$ in $\gG$}
    \end{aligned}
\end{equation}

\begin{example}
\label{ex:example-graph}
    Consider a toy Twitter-bot detection classifier that takes as input the \emph{days since the
    account was created}, and the \emph{total number of replies to the tweets made by this account},
    and outputs a binary decision: bot or not.  Starting from an arbitrary account, the adversary
    wants to create a bot that evades the detector by only modifying these two features.  To save
    time and money the adversary wants to keep these modifications to a minimum.

    In this setting, the transformation graph can be built as follows. For each feature vector $v
    \in V$ representing an account, there exist up to four children in the graph: an example with
    the value of the \emph{number of days since account creation} feature incremented by one, or
    decremented by one, and analogously two children for the \emph{number of replies to the tweets}.
    Let all edges have cost 1. In such a graph, the cost of a transformation chain is the number of
    edges traversed, e.g., incrementing the \emph{number of days since account creation} by three is
    equivalent to a path with three edges (the path cost is 3). The adversary's goal is to find the
    path with the lowest cost (minimal number of transformations) that flips the classifier's
    decision. The resulting account is the solution to~\Eqref{eq:mincost-graph-problem}.

    % Using the mimicry approach, the adversary wants to find a sequence of changes that flip the
    % decision of the classifier. Each of these changes incur a constant cost for the adversary, hence
    % it wants to use the minimal number of changes. The resulting modified account is the solution
    % to~\Eqref{eq:mincost-graph-problem}.
\end{example}

% Note that even though the cost between two immediately neighbouring examples $x$ and $x'$ is
% $\edgecost(x, x')$, the overall cost that is minimized in Eqn.~\ref{eq:mincost-graph-problem} is
% \emph{not} $\edgecost(x, \optim{x})$. The quantity being minimized is the path cost, that is the
% sum of edge costs $\edgecost(x, x')$ between all consecutive transformations $x$, $x'$ on the
% path. Thus, the construction of the graph allows to define various cost functions. \todo{Refer to
% the example graphs.}

\subsection{Provable Optimality Guarantees}
For a given adversary model, and an initial example $x$, we define $\optim{c}$ as the minimal
cost of the transformations needed to achieve a misclassification goal:
\[\optim{c} = \min_{x' \in V} \graphcost(x, x') \text{ s.t. } \isgoal(x') = \top\]

The minimal $\optim{c}$, or a tight \emph{lower} bound on $\optim{c}$, for a given $x$ is a measure
of adversarial robustness of the model, equivalent to the notion of \newterm{pointwise adversarial
robustness}~\cite{BastaniILVNC16, FawziMF16}, and \newterm{minimal adversarial cost}
(MAC)~\cite{LowdM05}.

The MAC can be used to quantify the \newterm{security of models}: the more
secure a model is, the higher the cost of successfully mounting an evasion attack.
In \Secref{sec:eval-bots} we illustrate this idea in the context of an ML classifier
for Twitter-bot detection, usign $\optim{c}$ to evaluate the security.

Finding the globally optimal $\optim{c}$ could be computationally expensive. A tight \emph{upper}
bound on $\optim{c}$, however, is easier to find in practice. In \Secref{sec:eval-wfp} we use upper
bounds on $\optim{c}$ to evaluate the effectiveness of evasion attacks as privacy defenses against
traffic analysis.

% !TEX root = ../oakland/main.tex

\section{Provably Minimal-Cost Attacks Using Heuristic Graph Search}
\label{sec:optim-attack}

One way to find an optimal, or admissible, solution to the graph-search problem
in~\Eqref{eq:mincost-graph-problem} is to use uniform-cost search (see \Secref{sec:model}). This
approach, however, can be inefficient or even infeasible. For example, let us consider the
transformation graph in Example~\ref{ex:example-graph}, where the branching factor is 4. Assuming
that at most 30 decrements or increments can be performed to any of the features, the number of
nodes in this graph is bounded by $n = 4^{30} = 2^{60}$. Given that uniform-cost search (UCS) needs
to expand $n$ nodes in the worst case, if a single expansion takes a nanosecond, a full graph
traversal would take 36 years.

For certain settings, however, it is possible to use heuristics to identify the best direction in
which to traverse the graph, escaping the combinatorial explosion through the usage of heuristic
search algorithms like \astar (see \Secref{sec:bfs}). To ensure that these algorithms find the
admissible $\optim{x}$ it is sufficient that the heuristic is \emph{admissible}~\cite{DechterP85}:

\begin{definition}[Admissible heuristic]\label{def:admissibility}
    Let $\gG = (V, E, \edgecost)$ be a weighted directed graph with $\edgecost(v, v') \geq 0$. A
    heuristic $h(v)$ is admissible if for any $v \in V$ and any goal node $g \in V$ it never
    overestimates the $\graphcost(v, g)$: $h(v) \leq \graphcost(v, g)$.
\end{definition}

In general, admissibility does not guarantee that \astar runs in an optimally \emph{efficient} way.
To guarantee optimality in terms of efficiency the heuristic must be \newterm{consistent}, a
stronger property~\cite{DechterP85}.

\subsection{Optimal Instantiation}
We detail one setting for which there exists an admissible heuristic for the adversarial example
search problem. Let the input domain $\sX$ be a discrete subset of the vector space $\sR^m$, and let
the cost of an edge $(\vx, \vx')$ in the transformation graph be a norm-induced metric between
examples $\vx$ and $\vx'$:
\[\edgecost(\vx, \vx') = \norm{\vx - \vx'},\]

% Note that even though the \emph{edge} cost is a norm between two feature vectors, the adversarial
% costs are sum of edges along paths in the graph, and hence can encode more complex cost functions as
% we show in \Secref{sec:eval-bots}.

Let $\domainclosure \subseteq \sR^m$ be a superset of $\sX$, e.g., a continuous closure of a
discrete $\sX$. Let $\dist(\vx)$ denote the minimal adversarial cost of the classifier at input $\vx$ with
respect to cost $\norm{\vx - \vx'}$ over the search space $\domainclosure$. Because the search space is a
subset of $\sR^m$, $\dist$ can be simplified from
\Eqref{eq:mincost-problem} to the following:
\begin{equation}\label{eq:robustness}
    \dist(\vx) = \min_{\Delta \in \domainclosure} \norm{\Delta} %
    \text{ s.t. } \isgoal(\vx + \Delta) = \top,\ \vx + \Delta \in \domainclosure
\end{equation}

Any \newterm{lower bound} $\distlowbound(\vx)$ on $\dist(\vx)$ over any $\domainclosure$ such that
$\sX \subseteq \domainclosure$, can be used to construct an admissible heuristic $\hinputspace(\vx)$:
\begin{equation}\label{eq:heuristic-threshold}
    \hinputspace(\vx) = \begin{cases}
        \distlowbound(\vx), &\isgoal(\vx') \neq \top \\
        0, &\text{otherwise}
    \end{cases}
\end{equation}

If $\vx$ is not already classified as the target class $t$, this heuristic returns the lower bound
on the MAC, $\distlowbound(\vx)$. When $\vx$ is classified as the target class, i.e., $\vx$ is
already on the other side of the decision boundary, the heuristic returns $0$. The heuristic $h$ is
admissible because it returns a lower bound on the path cost from an example $\vx$ to any
adversarial example $\optim{\vx}$.

\begin{statement}[Admissibility of $\hinputspace$]\label{thm:admissibility}
    Let the transformation graph $\gG = (V, E, \edgecost)$ have $\edgecost(\va, \vb) =
    \norm{\va-\vb}$, and the initial example be $\vx \in V \subseteq \sR^m$. Then $\hinputspace$ is
    an admissible heuristic for the graph search problem from \Eqref{eq:mincost-graph-problem}.
    (Proof in \Appref{app:admissibility-proof})
\end{statement}

In the rest of the paper, we use $\dist(\vx)$, $\distlowbound(\vx)$ and ``heuristic'' interchangeably,
as $\dist(\vx)$ or $\distlowbound(\vx)$ unambiguously define the heuristic $\hinputspace(\vx)$ through
\Eqref{eq:heuristic-threshold}.

We note that in many domains, and in particular in non-image domains, the transformation cost for
different features might not be equally distributed. \Stmtref{thm:admissibility} holds for any norm.
Hence, the edge cost can be instantiated with weighted norms to capture differences in adversarial
cost between features. We note that, regardless of the cost model, the structure of the
transformation graph can encode more complex cost functions than $\lp$ distances between vectors, as
we demonstrate in \Secref{sec:eval-bots}.

\begin{figure}
    \centering
    \input{parts/extras/illustration.tex}
\end{figure}

\subsection{Computing the Heuristic}\label{sec:computing-heuristic}

\subsubsection{Linear Models}
The MAC $\dist(\vx)$ over a continuous $\sR^m$ is equivalent to the standard notion of pointwise
adversarial robustness, and can be computed efficiently for linear models.
%
% In terms of the graphical framework, there are two settings where this is applicable. First, when
% the edge cost is defined over the feature space: $\edgecost(a, b) = \norm{\phi(a) - \phi(b)}$, which
% makes the problem in~\Eqref{eq:mincost-problem} similar to the feature adversary attack
% \cite{SabourCFF15}.  Second, a special case of the first one, when the feature map is the identity
% transformation $\phi(\vx) = \vx$, that is, the model is a linear model over the input space.

% These two scenarios are equivalent, as in both cases the target classifier is a linear model.
%
When the target model is a linear model, $\phi(\vx) = \vx$, $\dist(\vx)$ is a distance
from $\vx$ to the hyperplane defined by the discriminant function
(see \Figref{fig:heuristic-illustration}), and has a closed form:
\begin{equation}\label{eq:dist-decision-boundary}
    \dist(\vx) = \frac{|\vw \cdot \vx + b|}{\dualnorm{\vw}} = \frac{|f(\vx)|}{\dualnorm{\vw}},
\end{equation}
where $\dualnorm{\vw}$ is the dual norm corresponding to $\norm{\vw}$
\cite{Mangasarian99,PlastriaCarrizosa01}.
% This generic result for any
% norm is due to \citet{Mangasarian99}, and, independently, \citet{PlastriaCarrizosa01}.
%
If the edge cost is induced by an $\lp$ norm, $\edgecost(\va, \vb) = \norm{\va - \vb}_p$, the
denominator $\dualnorm{\vw}$ is $\norm{\vw}_q$, where $q$ is the H\"older conjugate of $p$:
$\frac{1}{p} + \frac{1}{q} = 1$.

\subsubsection{Non-Linear Models}
For some models, $\dist(\vx)$ can be either computed using formal methods~\cite{CarliniKBD17,
KatzBDJK17, BastaniILVNC16}, or bounded analytically, yielding a lower bound
$\distlowbound(\vx)$~\cite{TsuzukuSS18, PeckRGS17, HeinA17}. Existing methods usually perform the
computation over a box-constrained $\domainclosure = I_1 \times I_2 \times \cdots \times I_m$ for
some contiguous intervals $I_j \subset \sR$, and are only applicable to $\lp$ norm-based costs.
These methods are much more expensive to compute than the closed-form solution for linear models
above.

\subsubsection{Bounded Relaxations}\label{sec:optim:bounded}
A number of works explore bounded relaxations of the admissibility properties of \astar search,
aiming to trade off admissibility guarantees for computational efficiency~\cite{Pohl70, Pohl73,
PearlK82, LikhachevGT03}. In this paper, we employ \emph{static weighting}~\cite{Pohl70} for its
simplicity. In this approach, the heuristic value is multiplied by $\varepsilon > 1$. This results
in adversarial examples that have at most $\varepsilon$ times higher cost than MAC.

% !TEX root = ../oakland/main.tex

\section{Experimental Evaluation}
\label{sec:eval}

We evaluate the graph search approach for finding adversarial examples as \emph{means to
evaluate the security of an ML model} by computing its robustness against adversarial examples, and
as \emph{means to build efficient defenses against privacy-invasive ML models}. We use two ML
applications that work with constrained discrete domains: a bot detector in a white-box setting,
where it is possible to use \astar with admissible heuristics to obtain provably minimal-cost
adversarial examples, and a traffic-analysis ML classifier in a black-box setting, where we obtain
upper bounds on the minimal adversarial cost.

\descr{Implementation}
We use scikit-learn~\cite{scikit-learn} for training and evaluation of ML models, and Jupyter
notebooks~\cite{jupyter} for visualizations.  The reported runtimes come from executions on a
machine with Intel i7-7700 CPU working at 3.60GHz.  The code to run the attacks is available as a
Python package; all experiments are reproducible\footnote{[Link to the code anonymized]}.

\nocite{gnu-parallel}

% !TEX root = ../oakland/main.tex

\subsection{Evaluating Security: Twitter-Bot Detection}
\label{sec:eval-bots}

In this section, we evaluate the security of an ML-based Twitter-bot detector.  First, we show how
to use the graphical framework to compute adversarial robustness as the minimal cost of building a
bot that can evade detection, and compare the guarantees the framework provides to the standard
adversarial robustness measures. Second, we evaluate the efficiency and optimality of our attacks.
At the end of this section, we discuss the implications of our assumptions when the framework is
used as a security evaluation tool.
% 
% In this section we evaluate the effectiveness of our method as a means to evaluate the security of
% an ML model by computing its robustness against adversarial examples.
% We consider the case of an ML-based Twitter bot detector, in which the modeler uses minimal-cost
% transformations to quantify the ability of the adversary to build bot accounts that can evade
% detection.

As in our toy example, we assume the adversary starts with a bot account and aims to find the
minimum transformation that make the model classify the account as human.  They define a
transformation graph such that any chain of transformations results in a feasible account, runs the
graph search to find a minimal-cost example, and replicates the transformations on their bot account
to evade the classifier.

Note that, as opposed to adversarial examples on images where the perturbations added by the
adversary may change the content of the image to the point where it changes its class, in our
setting, a bot account will keep being a bot account regardless of the transformations.

\subsubsection{Twitter-Bot Detector}

%\descr{Target classifier}
We use a linear model as the target classifier that classifies an account as a ``bot'' or ``not
bot''.  In particular, we use $\lp[2]$-regularized logistic regression, as the use of a linear
model enables us to compute the exact value of the heuristic efficiently (see
\Secref{sec:computing-heuristic}).  The decision threshold of the classifier is standard:
$\probthresh = 0.5$.
We use 5-fold cross-validation on the training set (see below) to pick the $\lp[2]$ regularization
parameter of the logistic regression (set to $1.9$). Although simple, this classifier yields an accuracy
of 88\% (random baseline is 65\%) and performs on par with an SVM with an RBF kernel, and better than
some neural network architectures (see \Secref{sec:bots-discussion}). Hence, we consider the
regression to be a realistic choice in our setting.

\descr{Dataset}
We use the dataset for Twitter bot classification by \citet{GilaniKC17}. Each example in this
dataset represents aggregated information about a Twitter account in April of 2016. Concretely, it
has the following features: the \emph{number of tweets,} \emph{retweets,} \emph{favourites,}
\emph{lists,} and \emph{replies,} the \emph{average number of URLs,} the \emph{size of attached
content,} \emph{average likes} and \emph{retweets per tweet,} and the \emph{list of apps that were
used to post tweets} (see \Tabref{tab:bots-features}).

Accounts are human-labeled as bots or humans. The original dataset is
split into several bands by the number of followers. We report the results for the 1,289
accounts with under 1,000 followers; more popular accounts result in similar behavior.
We randomly split the dataset into training and test sets, containing 1160 and 129 accounts or, respectively,
90\% and 10\% samples.
%We use these splits throughout most of our experiments.
We generate adversarial examples for the 41 accounts in the test set that are classified as bots by the target
classifier.

\begin{table}[t]
    \caption{Features from the Twitter bot classification dataset by \citet{GilaniKC17}}%
\label{tab:bots-features}
\newcommand{\featurename}[1]{\texttt{#1}}
{\centering
\resizebox{\columnwidth}{!}{
    \begin{tabular}{lll}
    \toprule
        \textbf{Feature}
        & \textbf{Column name}
        & \textbf{Type} \\
    \midrule
        Number of tweets                       & \featurename{user\_tweeted}              & Integer                   \\
        Number of retweets                     & \featurename{user\_retweeted}            & Integer                   \\
        Number of replies                      & \featurename{user\_replied}              & Integer                   \\
        Age of account, days                   & \featurename{age\_of\_account\_in\_days} & Integer                   \\
        Total number of URLs in tweets         & \featurename{urls\_count}                & Integer                   \\
        Number of favourites (normalized)      & \featurename{user\_favourited}           & Float                     \\
        Number of lists (normalized)           & \featurename{lists\_per\_user}           & Float                     \\
        Average number of likes per tweet      & \featurename{likes\_per\_tweet}          & Float                     \\
        Average number of retweets per tweet   & \featurename{retweets\_per\_tweet}       & Float                     \\
        Size of CDN content, kB                & \featurename{cdn\_content\_in\_kb}       & Float                     \\
        Apps used to post the tweets           & \featurename{source\_identity}           & Set                       \\
        Total number of apps used              & \featurename{sources\_count}             & Integer                   \\
        Followers-to-friends ratio$^\dagger$             & \featurename{follower\_friend\_ratio}    & Float \\
        Favourites-to-tweets ratio$^\dagger$             & \featurename{favourite\_tweet\_ratio}    & Float \\
        Average cumulative tweet frequency$^\dagger$     & \featurename{tweet\_frequency}           & Float \\
    \bottomrule
    \end{tabular}
}}
\begin{minipage}{\columnwidth}
\vspace{1em}
{\footnotesize $\dagger$ We have dropped this feature from the dataset, since the effect of
transformations on it cannot be computed without having original tweets. We were not able to
obtain the original dataset of the tweets for compliance reasons.}
\end{minipage}

\end{table}

\descr{Feature Processing}
Almost all the features in the dataset are numeric (e.g., \emph{size of attached content}). We use
quantile-based bucketization to distribute them into buckets that correspond to quantiles in the
training dataset. In our experiments, we use 20 buckets, which offers best performance in a grid
search measuring 5-fold cross-validation accuracy on the training set. After bucketization, we
one-hot encode the features.

The only non-numerical feature is the \emph{list of apps that were used to post the tweets}.  We
encode it as follows. For each of the six apps in the dataset, we use two bits: if the app was used
by the account we set the first bit, and if not, we set the second bit.

\subsubsection{Security Evaluation}
Here, we evaluate the security of the bot detector using minimal adversarial costs as the measure of
adversarial robustness.

\descr{The Adversary's Goals}
% The minimal adversarial cost provides insights into the minimal resources required for an adversary to fulfill her goal of fooling the detector.
As mentioned in \Secref{sec:model:goal}, the minimal adversarial costs depend on the adversary's
definition of ``fooling''. To illustrate how the framework can accommodate different goals, we
simulate two attack settings. First, a \emph{basic attack}, in which the adversary's goal is to find
any adversarial examples that flip the decision of the classifier ($\conflevel = \probthresh =
0.5$).  Second, a \emph{high-confidence attack}, in which the adversary's goal is to find
adversarial examples that are classified as ``not bot'' with at least \emph{75\% confidence}
($\conflevel = 0.75$).

\descr{The Adversary's Capabilities: Transformation Graph and Cost}
For this evaluation, we assume that the adversary is capable of changing all account features, and
the cost of an adversarial example is the \emph{number of changes} required to transform an initial
bot account into that example. This adversarial cost model is similar to the state of the art in
adversarial ML (see \Secref{sec:related-instantiations}). In \Secref{sec:bots-discussion}, we
discuss a different model in which the adversary is constrained by the number of features that they
can influence, and the cost is not measured in the number of changes, but in the actual dollar cost
of performing transformations.

We build the transformation graph by defining \emph{atomic} transformations that change \emph{only
one} feature value. For each bucketed feature we define two atomic transformations: increasing the
feature value so that it moves one bucket up, and decreasing the feature value so that it moves one
bucket down. For the buckets in the extremes, only one transformation is possible. For the \emph{list
of apps} feature, we define one transformation per app: flipping the bits that represent whether the
app was used or not. Then, all possible modifications of an initial example, including those that
change multiple features, can be represented as chains of atomic transformations: paths in the
graph. For example, a modification that changes two features needs at least two atomic
transformations: a path with two edges.

\begin{figure}
    \centering
    \resizebox{0.9\columnwidth}{!}{
        \centering
        \vspace{1em}
        \begin{tikzpicture}
        \tikzstyle{featurevec}=[draw, fill=lightgray!10, text width=5cm, font=\sffamily, scale=2]
        \tikzstyle{caption}=[font=\sffamily, scale=2]
        \node[featurevec] at (0,0) (a) {Number of tweets: few \\ Age of account: medium};
        \node[featurevec] at (6,-4) (b) {Number of tweets: \textit{medium} \\ Age of account: medium};
        \node[featurevec] at (-6,-4) (c) {Number of tweets: few \\ Age of account: \textit{new}};
        \node[caption] at (14,-2) (ellipsis1) {...};
        \node[featurevec] at (-5,-8) (d) {Number of tweets: \textit{medium} \\ Age of account: \textit{new}};
        \node[featurevec] at (7,-8) (e) {Number of tweets: \textit{medium} \\ Age of account:
        \textit{$>$
        1 year}};
        \draw[->] (a)->(b) node[caption] [left=3, midway] {2};
        \draw[->] (a)->(c) node[caption] [left=5, midway] {2};
        \draw[->, color=black!50] (a)->(ellipsis1);
        \draw[->] (b)->(d) node[caption] [left=15, midway] {2};
        \draw[->] (c)->(d) node[caption] [left=1, midway] {2};
        \draw[->] (b)->(e) node[caption] [left=1, midway] {2};
        \end{tikzpicture}
    }
\caption{
    Sketch of the transformation graph for simplified Twitter accounts. \emph{Italics} denote
    a feature value that differs from the initial example. The edge costs are constant and equal
    to the $\lp[1]$ distance between the feature vectors representing the accounts.}
\label{fig:bots-graph-illustration}
\end{figure}
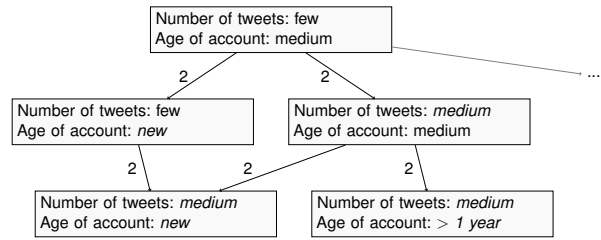

We define the edge costs to be the $\lp[1]$ distance between feature vectors before and after a
transformation. Given the one-hot encoding, this means that each atomic transformation in our graph
has a constant cost of 2 (one bit is set to zero, and another bit to one). Such a representation has
two advantages. First, the path cost can be easily related to the number of changes needed to evade
the classifier: it suffices to divide the path cost by two. Second, because the edge cost is a
norm-induced metric, we can use \astar with admissible heuristic by \Stmtref{thm:admissibility}.
\Figref{fig:bots-graph-illustration} illustrates an example transformation graph for simplified
accounts.

Note that the $\lp[1]$ distance between two arbitrary feature vectors does not represent the number
of changes between them. We are able to represent the number of changes through the structure of the
transformation graph.

% The edge costs in the graph in such a way that they satisfy two properties. First, as
% mentioned, in these experiments we want that path costs correspond to the number of feature changes,
% either of bucketized features, or the \emph{list of apps} feature.  Second, so that we can use
% \astar with admissible heuristic, they must meet the conditions of \Stmtref{thm:admissibility}---an
% edge cost $\edgecost(\vx, \vx')$ should be equal to $\norm{\vx - \vx'}$ for some norm.  Owing to the
% way the examples are represented, we can set the edge weights to $\lp[1]$ distance between feature
% vectors to satisfy both of the goals.  Each atomic transformation in our graph then has a constant
% cost of 2 (one bit is set to zero, and another bit to one). Hence, from the path cost of an
% adversarial example one can easily obtain the number of required changes to evade the
% classifier---by dividing the cost by two.  We discuss other possible choices of edge costs ($\lp[2]$,
% $\lp[\infty]$ distances) that would result in the same cost model shortly.

\descr{Results}
We run \astar search on the transformation graph to find adversarial examples. For the basic attack,
i.e., when the adversary wants to find an example that flips the decision regardless of the
confidence, on average only 2.2 (s.d. 1.6) feature changes suffice to flip the decision of the
target model. To obtain adversarial examples with high confidence ($75\%$), the adversary
needs on average 3.9 (s.d. 2.7) changes.
% These numbers are useful to rank different models against each other.
We report some examples of adversarial example accounts
in \Tabref{tab:bots-examples} (\Appref{app:figures}).

% For instance, they often change the \emph{list of apps} feature, removing the \emph{``other''}
% value, and adding \emph{``mobile''}. To enact this transformation, the adversary needs to make the
% tweets look as if they are coming from a mobile app.

\begin{figure*}[t]
        \centering
        \includegraphics[width=0.36\textwidth]{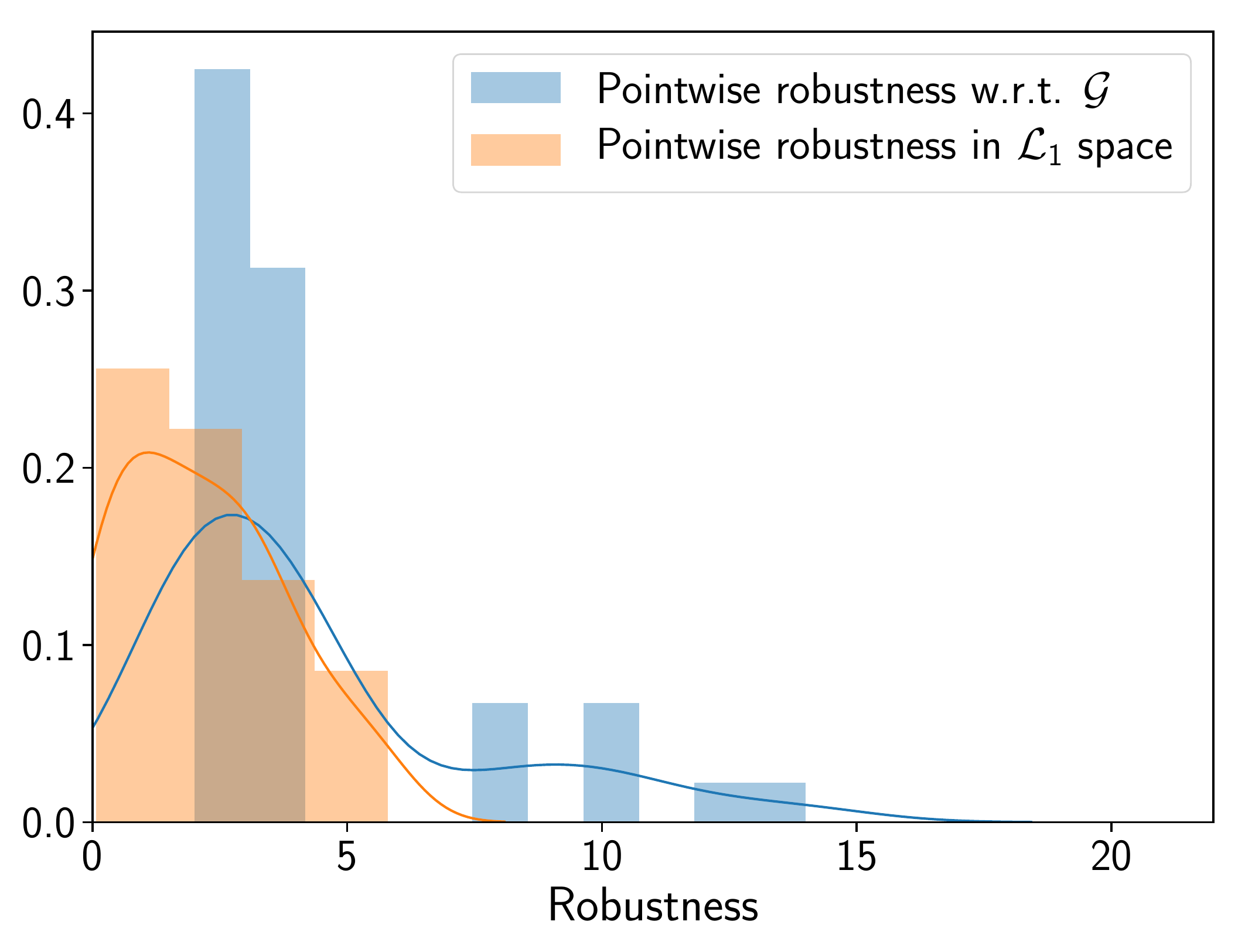}
        \includegraphics[width=0.36\textwidth]{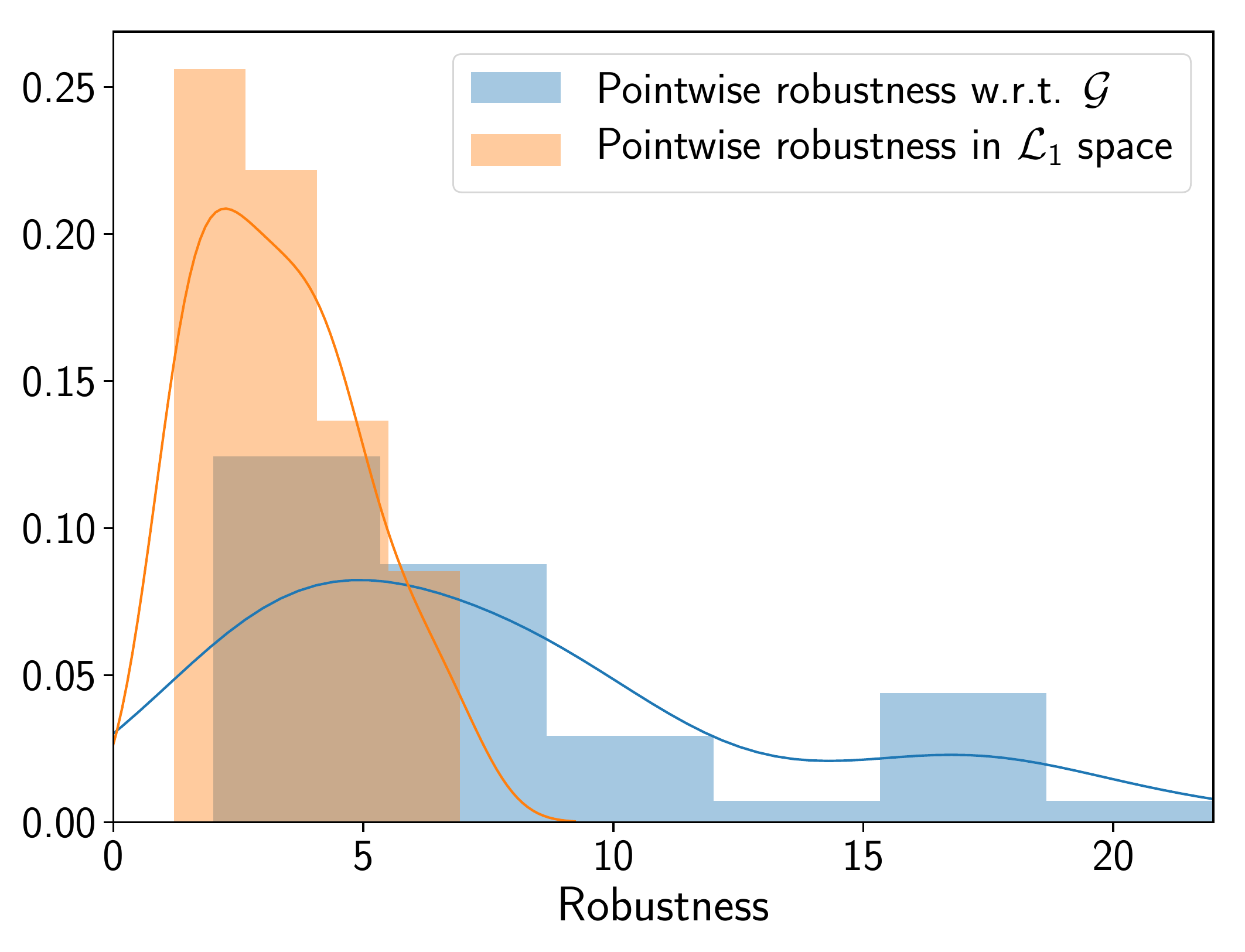}
    \caption{Pointwise adversarial robustness measured as minimal-cost
    adversarial examples computed using the transformation graph $\gG$, and pointwise robustness
    in $\lp[1]$ space. Left: basic attack. Right: high-confidence attack. (Notice the different
    y-axes)}
    \label{fig:bots-guarantees}
\end{figure*}

% \begin{figure*}[t]
%     \hspace{2em}
%     \begin{subfigure}[t]{0.364\textwidth}
%         \centering
%         \includegraphics[width=\textwidth]{images/bots__mac_distplots__band_1k__target_50__model_lr.pdf}
%         \caption{Basic attack}
%         \label{fig:bots-high-robustness-dist}
%     \end{subfigure}
%     \hspace{1em}
%     \begin{subfigure}[t]{0.364\textwidth}
%         \centering
%         \includegraphics[width=\textwidth]{images/bots__mac_distplots__band_1k__target_75__model_lr.pdf}
%         \caption{High-confidence attack}
%         \label{fig:bots-robustness-dist}
%     \end{subfigure}
%     \caption{Distribution of values of pointwise adversarial robustness obtained using minimal-cost
%     adversarial examples from search over the transformation graph $\gG$, and pointwise robustness
%     in $\lp[1]$ space.}
%     \label{fig:bots-guarantees}
% \end{figure*}

Next, we compare the minimal adversarial cost of adversarial examples as a security measure to the
standard notion of adversarial robustness over continuous $\domainclosure = \sR^m$ (as in
\Eqref{eq:robustness}).  We show in \Figref{fig:bots-guarantees} the distribution of the values of
both measures for the basic and high-confidence attacks. We see that the MAC values from our method
are up to 26$\times$ higher than adversarial robustness over the unconstrained $\lp[1]$ space in the
case of the basic attack, and up to 486$\times$ higher in the case of the high-confidence attack.
% Robustness in $\lp[1]$ space may be a small number, e.g., under 1, while a value of MAC
% from our transformation graph is necessarily a multiple of 2, since that is the cost of a single
% transformation---hence the ratio is this high.

This means that the continuous domain robustness measure applied to a discrete domain results in
overly pessimistic adversarial cost estimates, that an adversary \emph{cannot} achieve because of
inherent domain constraints. Our approach produces a more precise robustness measure, tailored to
the concrete domain constraints and the adversary's capabilities.

\subsubsection{Performance Evaluation}
% In the previous section, we evaluated the suitability of MAC as a useful measure of adversarial
% robustness.
Here, we study the trade-off between being able to run the graph search efficiently and the
optimality guarantees of the obtained MAC values.
% \descr{Search algorithms}
We consider the following algorithms: uniform-cost search (UCS), plain \astar, and
$\varepsilon$-bounded relaxations of \astar with $\varepsilon=\{2,3,5,10\}$ (see
\Secref{sec:optim:bounded}).

\begin{figure*}
    \centering
    % \begin{subfigure}[t]{0.36\textwidth}
    %     \centering
        \includegraphics[width=0.36\textwidth]{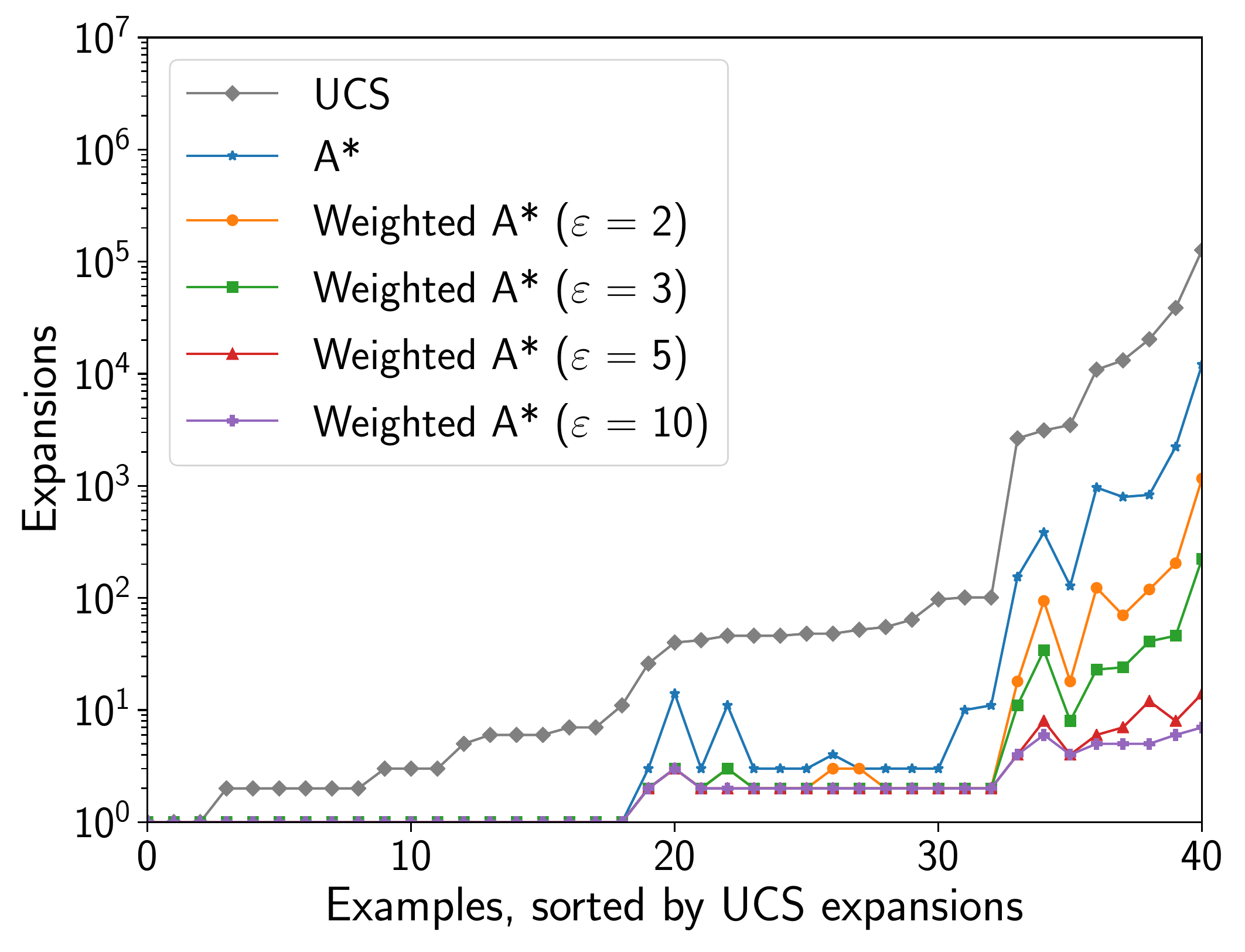}\quad
        % \caption{Graph node expansions}
    %     \label{fig:bots-expansions}
    % \end{subfigure}
    % \hspace{1em}
    % \begin{subfigure}[t]{0.36\textwidth}
    %     \centering
        \includegraphics[width=0.36\textwidth]{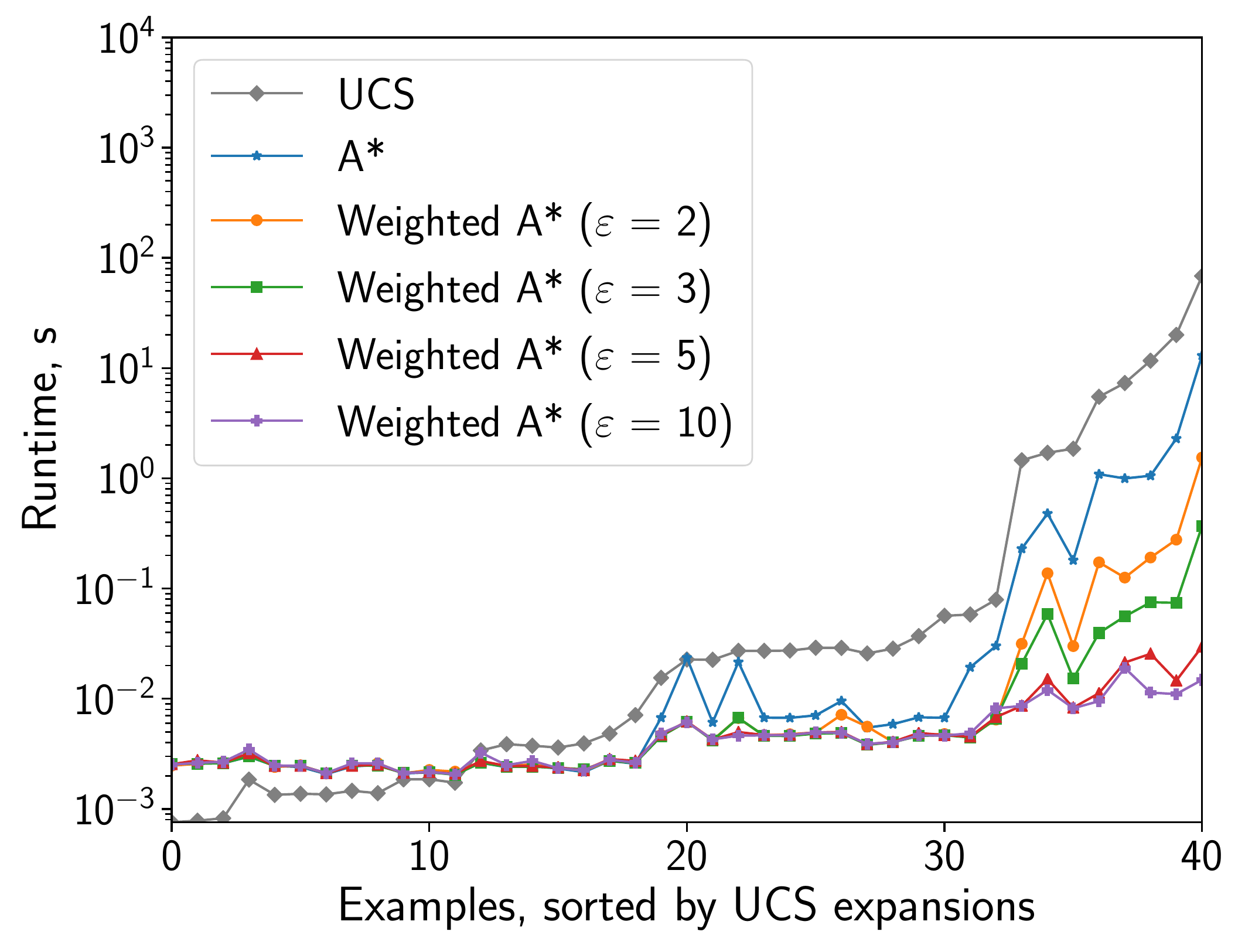}
        % \caption{Runtime}
        % |\label{fig:bots-runtimes}
    % \end{subfigure}
    \caption{Basic attack performance performance. Left: node expansions. Right: runtime in seconds. (y-axes are logarithmic)}
    \label{fig:bots-perf}
\end{figure*}

\descr{Runtime}
\Figref{fig:bots-perf} shows the number of expansions (left), as well as the runtime (right), needed
to find adversarial examples that flip the detector in the basic attack. We find that \astar expands
significantly fewer nodes than UCS (up to 32$\times$ fewer), showing that our admissible heuristic
is indeed useful for efficiently finding MAC adversarial examples in the search space. Relaxing the
optimality requirement by increasing the $\varepsilon$ weight speeds up the search even more. For
instance, $\varepsilon = 5$ decreases runtime by three orders of magnitude, and still ensures that
the minimal cost is at most five times lower than the cost of the found adversarial example.  We
also observe that in some cases UCS performs better than \astar in terms of the runtime, even
though \astar expands fewer nodes. We believe that this is an artifact of our Python implementation,
which could be solved with a more efficient implementation.

\begin{figure*}
    \centering
    % \begin{subfigure}[t]{0.365\textwidth}
    %     \centering
        \includegraphics[width=0.36\textwidth]{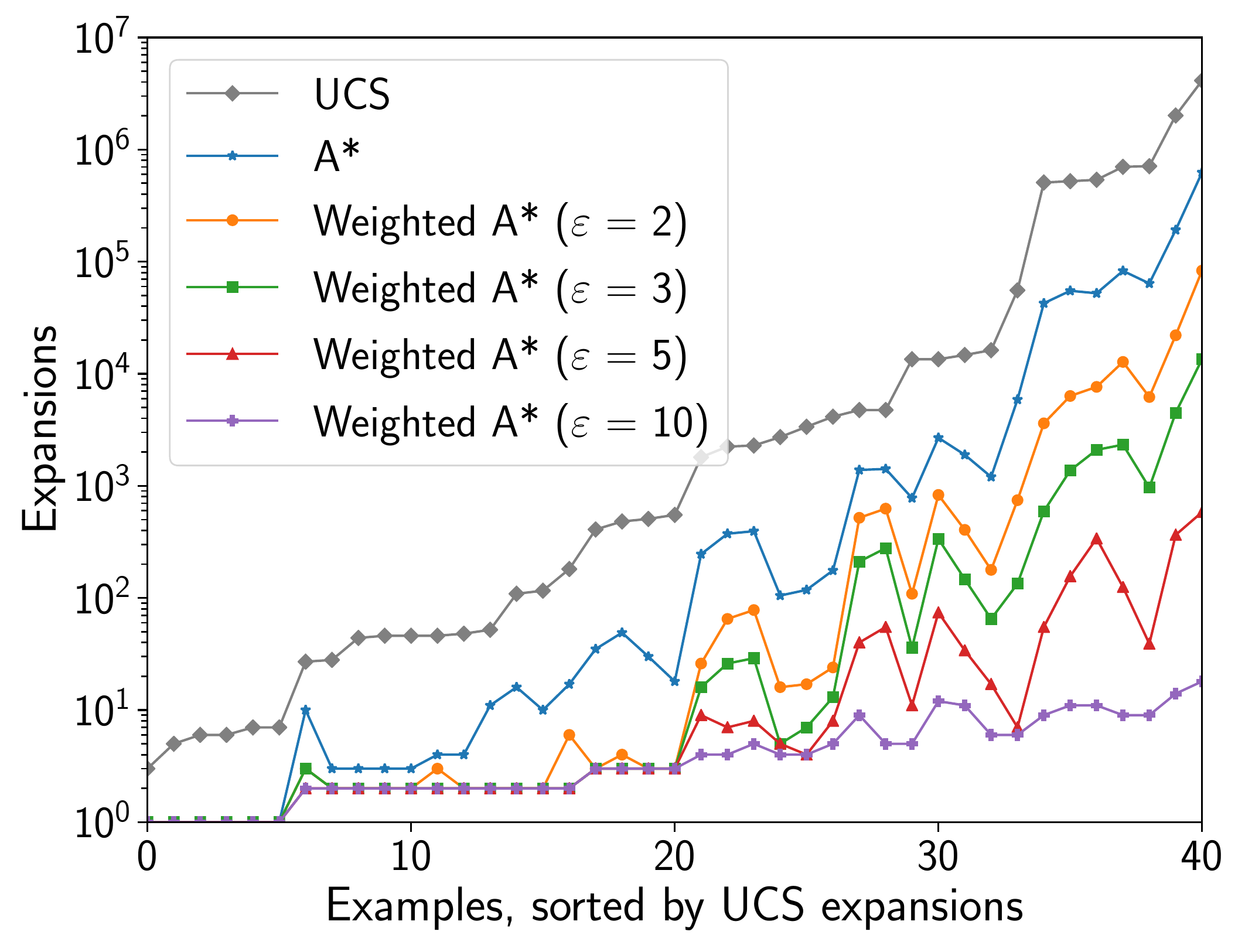}\quad
    %     % \caption{Graph node expansions}
    %     \label{fig:bots-high-expansions}
    % \end{subfigure}
    % \hspace{1em}
    % \begin{subfigure}[t]{0.365\textwidth}
    %     \centering
        \includegraphics[width=0.36\textwidth]{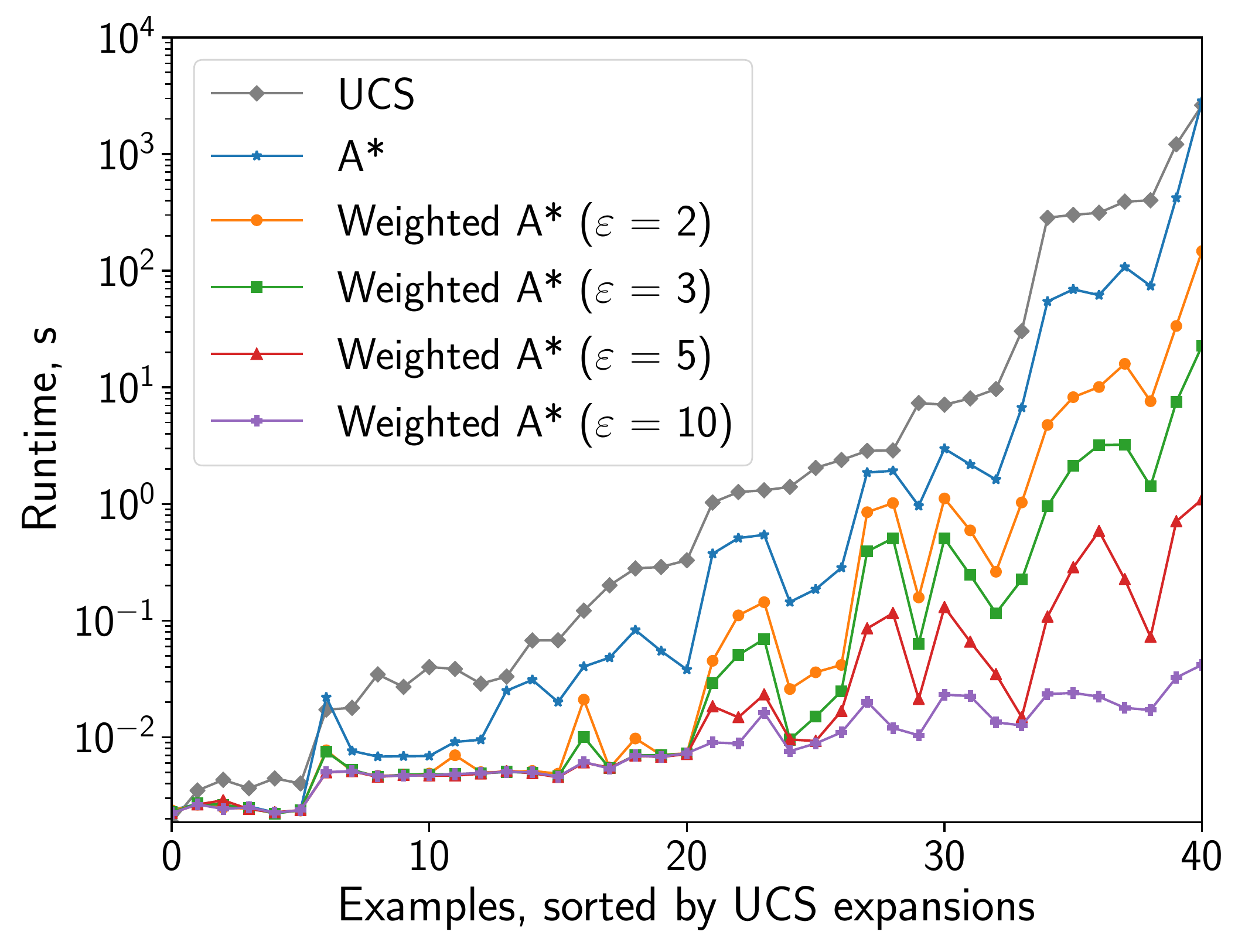}
    %     % \caption{Runtime}
    %     \label{fig:bots-high-runtimes}
    % \end{subfigure}
    \caption{High-confidence attack performance. Left: node expansions. Right: runtime in seconds. (y-axes are logarithmic)}
    \label{fig:bots-high-perf}
\end{figure*}

We observe similar results for the high-confidence attack (\Figref{fig:bots-high-perf}).
High-confidence adversarial examples require more transformations, hence, the search takes up to
100$\times$ more runtime than for the basic attack.  Still, \astar performs significantly better
than UCS, expanding 2--31$\times$ fewer nodes.

\begin{figure}[h]
    \centering
    \includegraphics[width=0.36\textwidth]{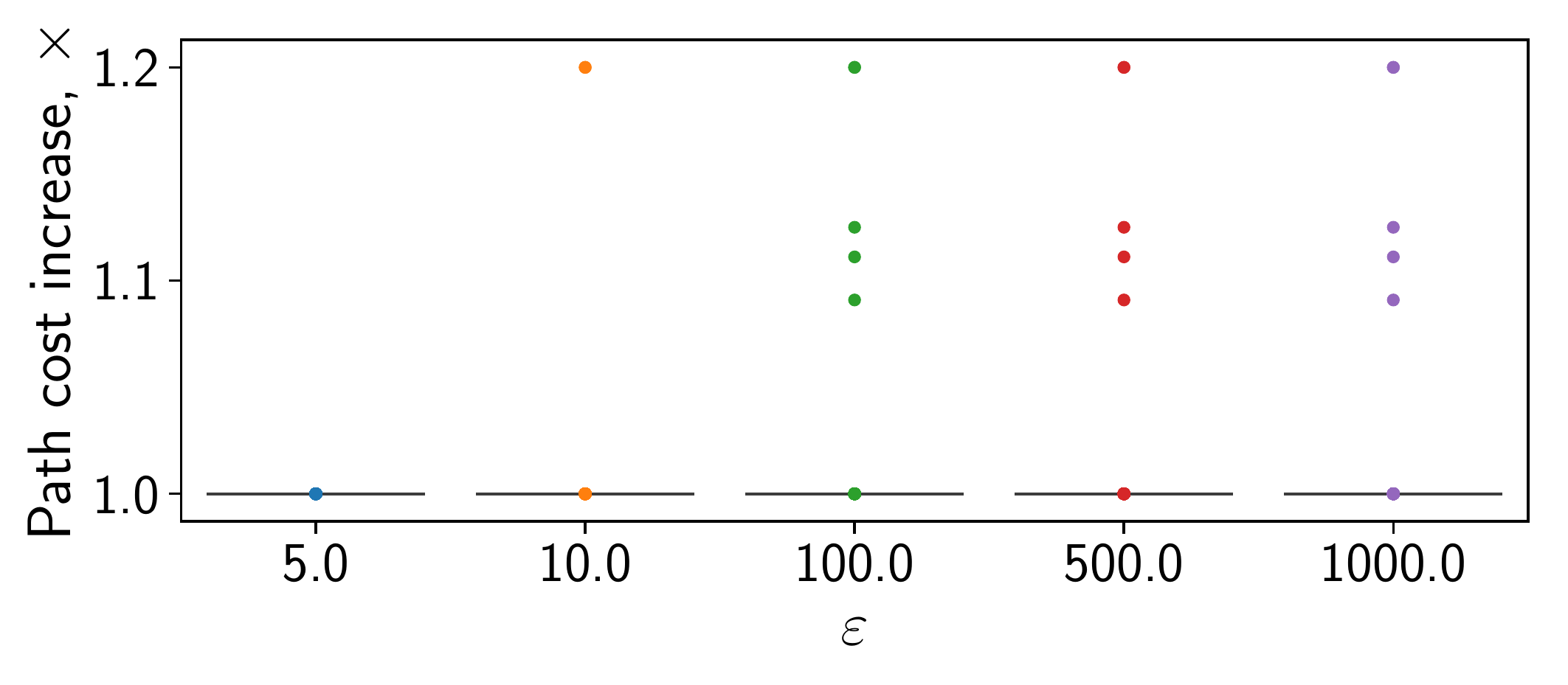}
    \caption{High-confidence attack: increase in cost over MAC of adversarial examples found
    using $\varepsilon$-weighted \astar.}
    \label{fig:bots-high-overhead}
\end{figure}

\descr{Speed vs. Optimality Trade-Off}
We saw that $\varepsilon$-bounded relaxations can drastically decrase the search runtime. We
empirically assess by how much this speedup hurts the optimality of obtained adversarial examples.
To evaluate this, we compute the increase in costs of adversarial examples found with
$\varepsilon$-weighted \astar over the optimal adversarial examples found with plain \astar. We
discover that the upper bound on sub-optimality from $\varepsilon$-weighted \astar is extremely
pessimistic (recall that $\varepsilon$-weighting guarantees that the found adversarial examples
have at most $\varepsilon$ times higher cost than the MAC). In practice, for the tested values of
$\varepsilon$, \emph{all} adversarial examples found in the basic attack incur minimal cost in our
transformation graph, and only few high-confidence adversarial examples have costs at most
$1.2\times$ higher than the MAC (see \Figref{fig:bots-high-overhead}).  We conclude that for this
setting, using $\varepsilon$-weighting can bring huge performance benefits at no cost.

\descr{Heuristics Comparison}
Up to this point, we used $\lp[1]$ distance as an edge weight, and the corresponding $\lp[1]$-based
heuristic in the search. Here, we investigate if other heuristics provide better performance.  To
maintain the optimality of \astar, the edge weights have to change accordingly. Given that we
consider that all transformations have the same cost, a uniform change in the weights results in an
equivalent transformation graph. The difference lies in the multiplicative factor used to recover
the number of required changes from the path cost.

We run the basic attack using $\lp[1]$, $\lp[2]$, and $\lp[\infty]$ edge weights, i.e., 2 for
$\lp[1]$, $\sqrt{2}$ for $\lp[2]$, and $1$ for $\lp[\infty]$; and the corresponding heuristics, that
we denote as $\dist[1], \dist[2], \dist[\infty]$. We compare the performance of \astar for these
edge costs against each other and against two baselines. First, UCS, which expands the same number
of graph nodes for all three edge cost models. It represents the worst case in terms of performance,
but outputs provably minimal-cost adversarial examples. Second, we run ``random search'' 10
times. By random search we mean \astar with a heuristic that outputs random numbers between 0 and 2
on the transformation graph with $\lp[1]$ edge costs. This algorithm is efficient, but does not
provide any optimality guarantees.

\begin{figure*}[h!]
    \centering
    % \hspace{2.4em}
    % \begin{subfigure}[t]{0.364\textwidth}
    %     \centering
        \includegraphics[width=0.36\textwidth]{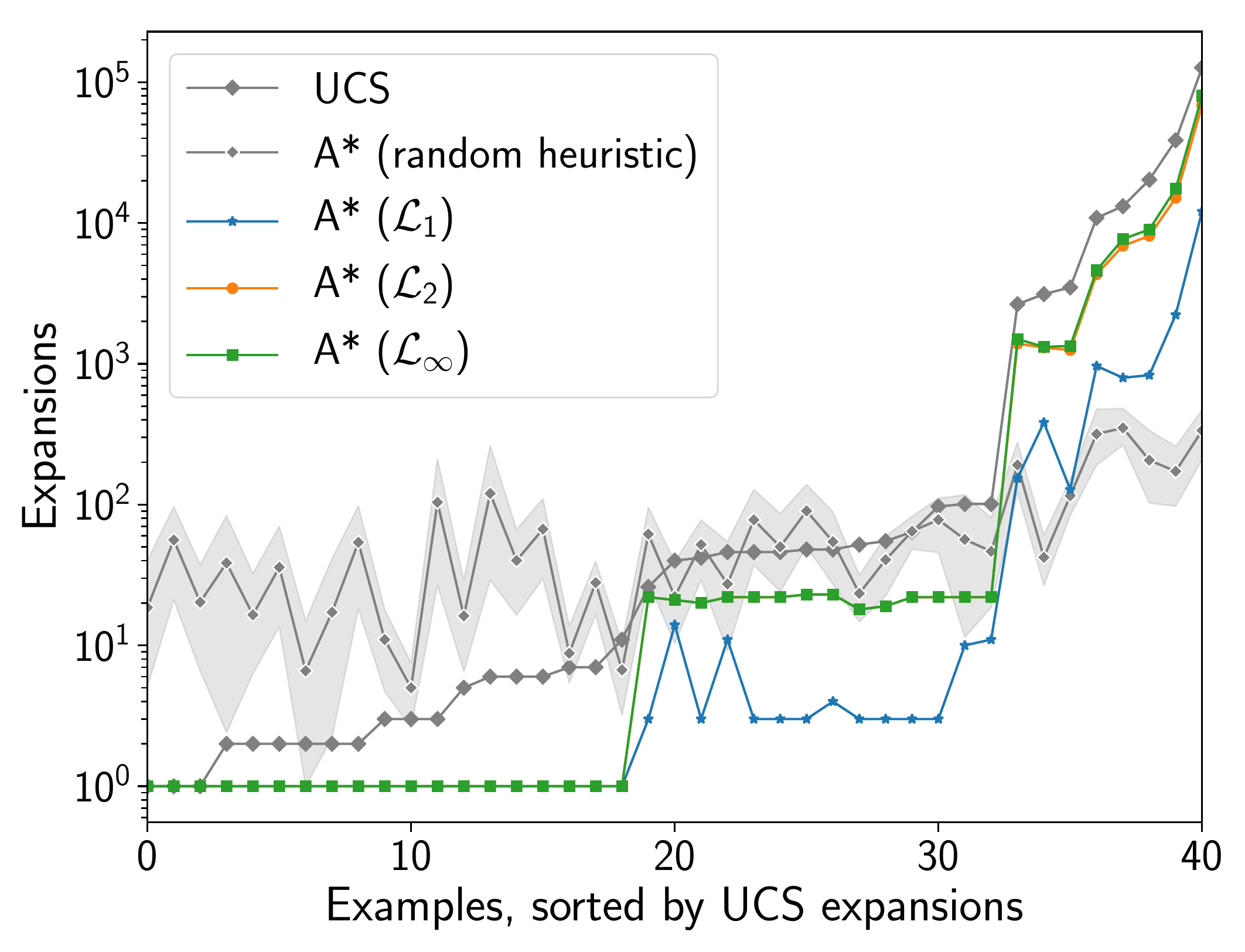}\quad
        % \caption{Graph node expansions}
    % \end{subfigure}
    % \begin{subfigure}[t]{0.364\textwidth}
        % \centering
        \includegraphics[width=0.36\textwidth]{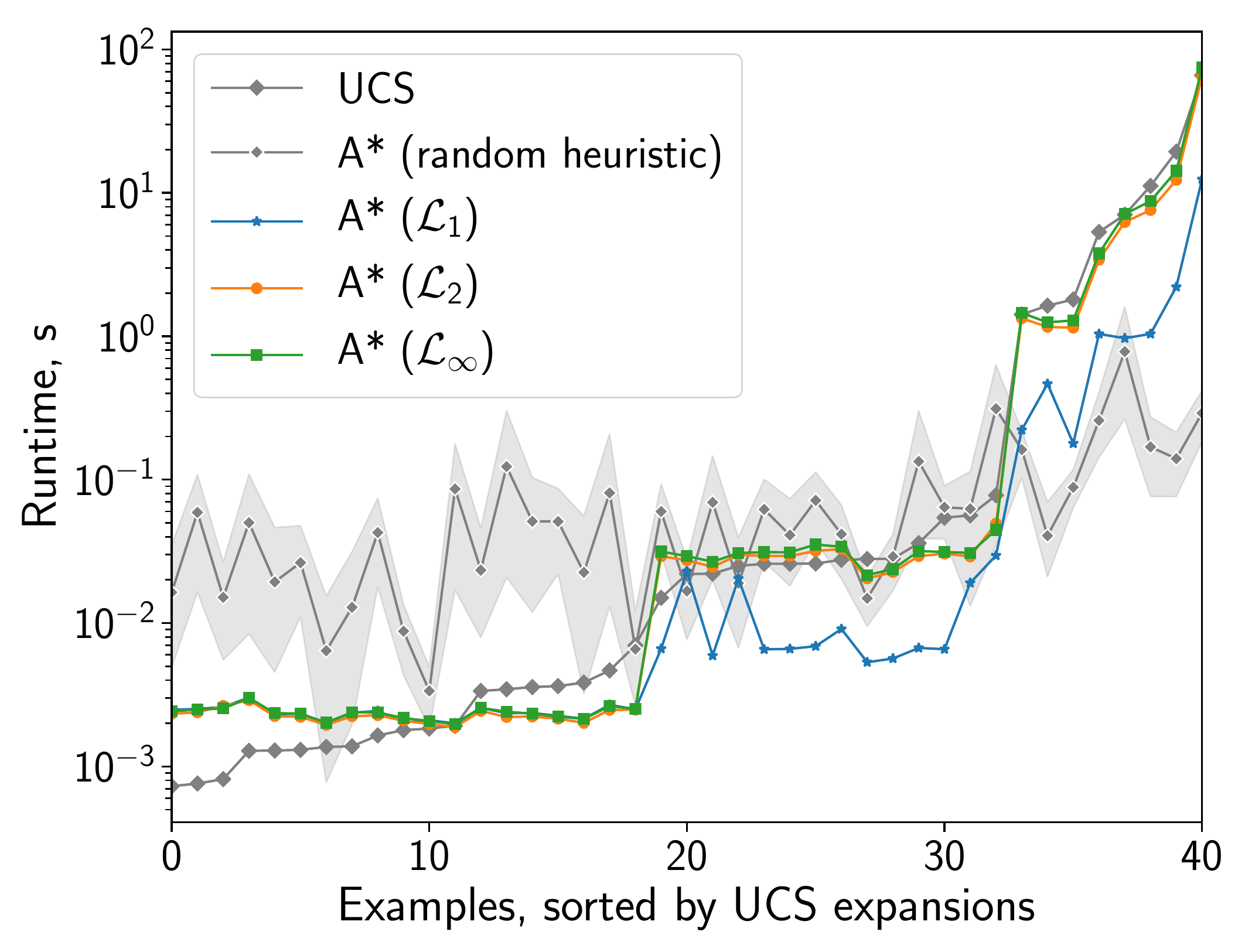}
        % \caption{Runtime}
    % \end{subfigure}
    \caption{Basic attack heuristics comparison. Left: node expansions. Right: runtime in seconds. (y-axes are logarithmic)}
    \label{fig:bots-heuristic-comparison}
\end{figure*}

We show the result in \Figref{fig:bots-heuristic-comparison} (left). Except for adversarial examples
that lie deeper in the graph, all admissible heuristics find solutions faster than the random
search. We also see that the random search outputs adversarial examples that have costs up to 11 times
the MAC obtained with \astar.

Even though both $\lp[2]$ and $\lp[\infty]$ heuristics enable the algorithm to explore fewer graph
nodes than UCS, in practice they take the same time (see \Figref{fig:bots-heuristic-comparison},
right) This is because each heuristic exploration is quite costly in terms of computation time. The
edge weights for $\lp[2]$ and $\lp[\infty]$ ($\sqrt{2}$ and $1$, respectively) are comparatively
high compared to the values of the heuristics (on our dataset $\dist[2]$ is on average 0.52 and
$\dist[\infty]$ is on average 0.059). Hence, \astar often needs to explore all nodes of the same
cost before it can proceed to transformations that carry a higher cost, essentially degenerating
into UCS. On the contrary, $\lp[1]$ heuristics perform consistently better, as $\dist[1]$ is on average 2.18,
higher than 2 (the $\lp[1]$ edge cost). We note that this result may not necessarily hold for other
transformation graphs with different cost models.

% values of the heuristic $\dist[1]$ (as an example, on
% our test dataset its average value is 2.18) are much higher than values of $\dist[2]$ (average 0.52
% on the test dataset) and $\dist[\infty]$ (average 0.059 on the test dataset). The edge costs,
% $\sqrt{2}$ and $1$, are comparatively higher than values of $\dist[2]$ and $\dist[\infty]$
% heuristics.  As a result, heuristic values are rarely higher than costs of applying further
% transformations.  \astar hence often needs to traverse all nodes of the same cost before moving on
% to nodes that have more transformations, degenerating into UCS. Having higher values, $\dist[1]$
% works better with the edge costs in our graph. This result does not necessarily hold for other
% transformation graphs with different cost models.

% \begin{figure}
%     \centering
%     \includegraphics[width=0.365\textwidth]{images/bots__random_overhead__20_band_1k__target_50.pdf}
%     \caption{Increase of path costs of adversarial examples found using random search over
%     minimal costs.}
%     \label{fig:bots-random-overhead}
% \end{figure}

\subsubsection{Applicability Discussion}
\label{sec:bots-discussion}
In the previous experiments, we assumed that the target model is linear, that the adversarial cost
is proportional to the number of feature changes, and that the adversary has white-box knowledge and
uses optimal algorithms. In this section, we explore other options for each of these assumptions and, in
\Secref{sec:eval-wfp}, we conduct a thorough evaluation of a setting in which none of them hold.

\descr{Non-Linear Models} When the target model is linear, we can efficiently compute the exact
value of the the admissible heuristic from \Eqref{eq:heuristic-threshold}. Even though a linear
model is a sensible choice in our setting, non-linear models can often appear in other
security-critical settings. Mounting an \astar-based attack against a
non-linear target model, however, requires costly methods to compute the heuristic (see
\Secref{sec:computing-heuristic}).

One way to overcome this issue is linearizing the non-linear model using a first-order Taylor
expansion, and then using the heuristic $\dist$ for linear models (see
\Appref{sec:non-linear-heuristic-deriv} for a formal derivation):
\begin{equation}\label{eq:dist-nonlinear-decision-boundary}
    \dist(\vx) \approx \frac{|f(\vx)|}{\dualnorm{\nabla_{\vx} f(\vx)}}
\end{equation}
We note that, as this approximation can overestimate $\dist(\vx)$, it cannot serve as an admissible
heuristic without additional assumptions on $f$.

\begin{figure*}
    \centering
    % \begin{subfigure}[b]{0.364\textwidth}
    %     \centering
        \includegraphics[width=0.36\textwidth]{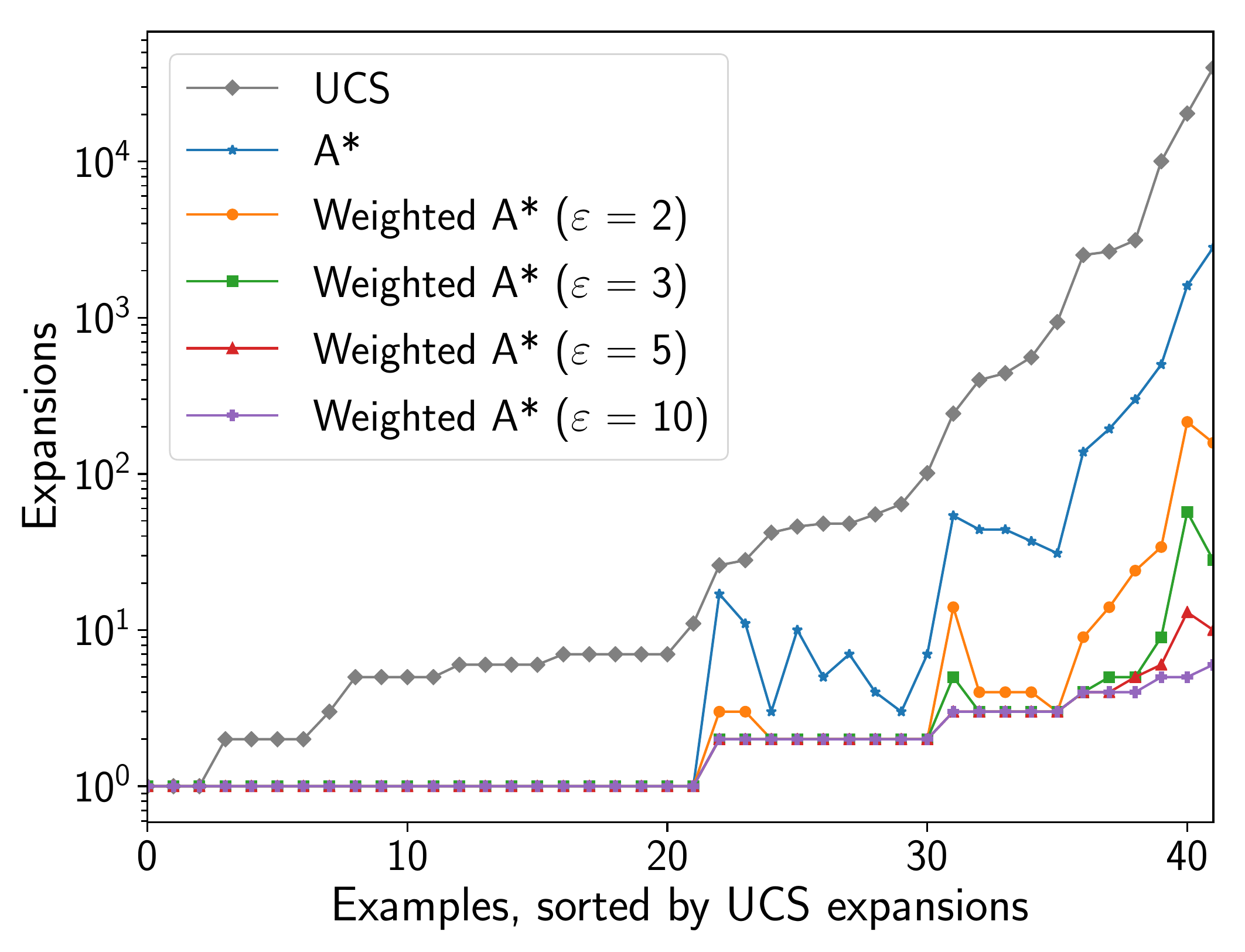}\quad
        % \caption{Graph nodes expansions}
        % \label{fig:bots-svmrbf-expansions}
    % \end{subfigure}
    % % \hspace{1em}
    % \begin{subfigure}[b]{0.364\textwidth}
    %     \centering
        \includegraphics[width=0.36\textwidth]{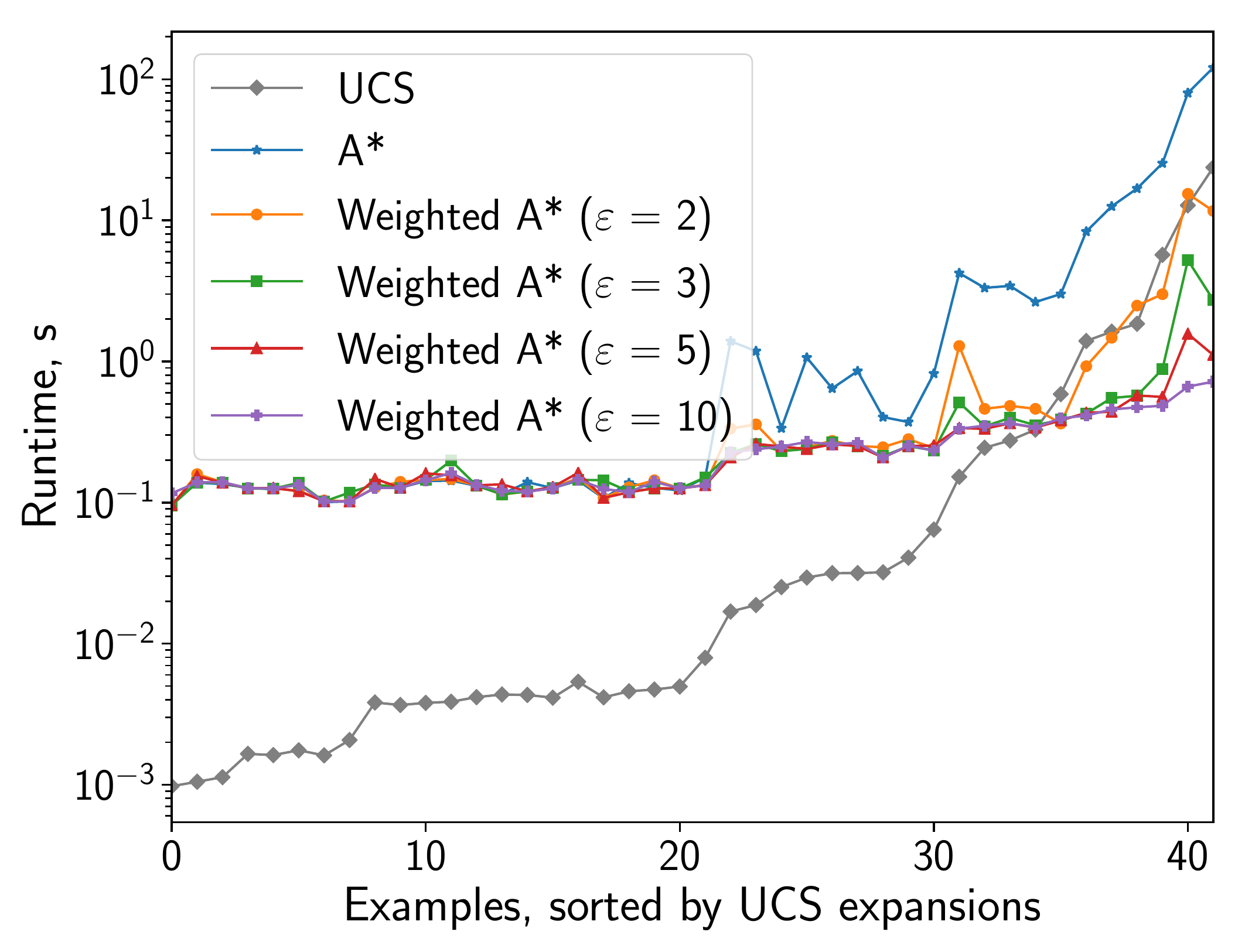}
        % \caption{Runtime}
    %     % \label{fig:bots-svmrbf-runtimes}
    % \end{subfigure}
    \caption{Basic attack against a non-linear model (SVM-RBF) using an
    approximate heuristic. Left: node expansions. Right: runtime in seconds. (y-axes are
logarithmic)}
    \label{fig:bots-svmrbf}
\end{figure*}

We empirically evaluate this heuristic by running the attack against an SVM with the RBF kernel
trained on the dataset with discretization parameter set to 20 (88\% accuracy on the test set).
Using UCS to obtain the ground-truth minimal-cost adversarial examples, we find that, even though
the heuristic is approximate, all adversarial examples found with \astar are minimal-cost.
Moreover, this heuristic allows the graph search to use significantly fewer graph node expansions
than UCS (\Figref{fig:bots-svmrbf}, left). On the downside, even though the algorithm expands fewer
nodes, the overhead of computing the heuristic (which includes computing the forward gradient of the
SVM-RBF) is high enough that there is no actual improvement in performance, unless we use
$\varepsilon > 2$ weighting (\Figref{fig:bots-svmrbf}, right). More efficient implementations of the
heuristic can result in better gains.

\begin{table}
    \caption{Transferability of minimal-cost adversarial examples from logistic regression
    to other models. Columns: model; \emph{Accuracy}---model's accuracy on the test set;
    \emph{Trans. (basic)}---transferability rate of adv. examples aiming to be missclassified
    with 50\% confidence to this model; \emph{Trans.  (high)}---same, with 75\% confidence.}
    \label{tab:bots-transferability}

    \centering
    \resizebox{\columnwidth}{!}{
    \begin{tabular}{lrrr}
        \toprule
            \textbf{Model} & \textbf{Accuracy, \%} & \textbf{Trans. (basic), \%} & \textbf{Trans. (high),
            \%} \\
        \midrule
        LR      &        88 &          ---   &   --- \\
        NN-A    &        80 &          49 &   84 \\
        NN-B    &        83 &          38 &   95 \\
        GBDT    &        87 &          74 &   95 \\
        SVM-RBF &        88 &          73 &   100 \\
        % LR      &        87.60 &          ---   &   --- \\
        % NN-A    &        79.84 &          48.72 &   83.72 \\
        % NN-B    &        82.95 &          37.84 &   94.87 \\
        % GBDT    &        86.82 &          74.36 &   95.12 \\
        % SVM-RBF &        87.60 &          72.50 &   100 \\
        \bottomrule
    \end{tabular}
    }
\end{table}

\descr{Black-Box Setting} The minimal-cost adversarial examples, and the robustness guarantees they
induce, are specific to a particular target model. Do other models misclassify these examples as
well? If yes, the attack would not only be effective in the white-box setting, but also in the
black-box setting, where the adversary does not know the exact architecture and weights of the
target model. Furthermore, there would be no need to use expensive heuristics for non-linear models.

% CT -- this definition seems obvious, do we need it?
% For an initial bot account $x$, we say that its corresponding adversarial example $\optim{x}$ found
% using \astar against the target classifier $F$ \newterm{transfers} to another classifier $F'$ if
% initially $F(x) = F'(x) = \text{``bot''}$, and $F'(\optim{x}) = \text{``not bot''}$.

We check whether the adversarial examples found in the previous section using \astar against a
logistic regression (LR) are misclassified by other non-linear models trained on the same dataset.
We choose four ML models representative of typical architectures: two instantiations of a two-layer
fully-connected ReLU neural network, one with 2000 and 500 neurons (NN-A), and one with 20 and 10
neurons in the respective layers (NN-B); gradient-boosted decision tree (GBDT); and an SVM-RBF. We
do not run extensive hyperparameter search to obtain the best possible performance of these models,
but we ensure that all of them have accuracy greater than $80\%$ (the random baseline is $65\%$). 

Table~\ref{tab:bots-transferability} shows the results of the experiment. We see that the basic
minimal-cost adversarial examples (which mostly have confidence only slightly higher than 50\%)
transfer to the other models in at least 38\% of the cases and in about 73\% of the cases for GBDT
and SVM-RBF. In the high-confidence setting more than 83\% of the adversarial examples transfer to
all models. We conjecture that when the goal is to find any adversarial examples, the minimal-cost
adversarial examples often exploit a weakness found only in their target model, and, hence, rarely
transfer. When the target confidence level is higher, the adversarial examples require more
transformations and become similar to non-bots as seen in the training data. Hence, they are more
likely to generalize.

\descr{Non-Optimal Algorithms}
So far we have considered that the adversary uses algorithms that provide optimality guarantees:
UCS, \astar, $\varepsilon$-weighted \astar. These algorithms are often expensive. We investigate
the performance of less expensive non-optimal algorithms in our setting using a hill-climbing
modification of \astar as an example (see \Secref{sec:bfs}). Regular \astar needs to keep track of
all previously expanded nodes at any given time. The hill-climbing variation only keeps the
best-scoring node. This significantly improves memory and computation requirements, but sacrifices
the ability of \astar to backtrack.

\begin{figure*}
    \centering
        \includegraphics[width=0.36\textwidth]{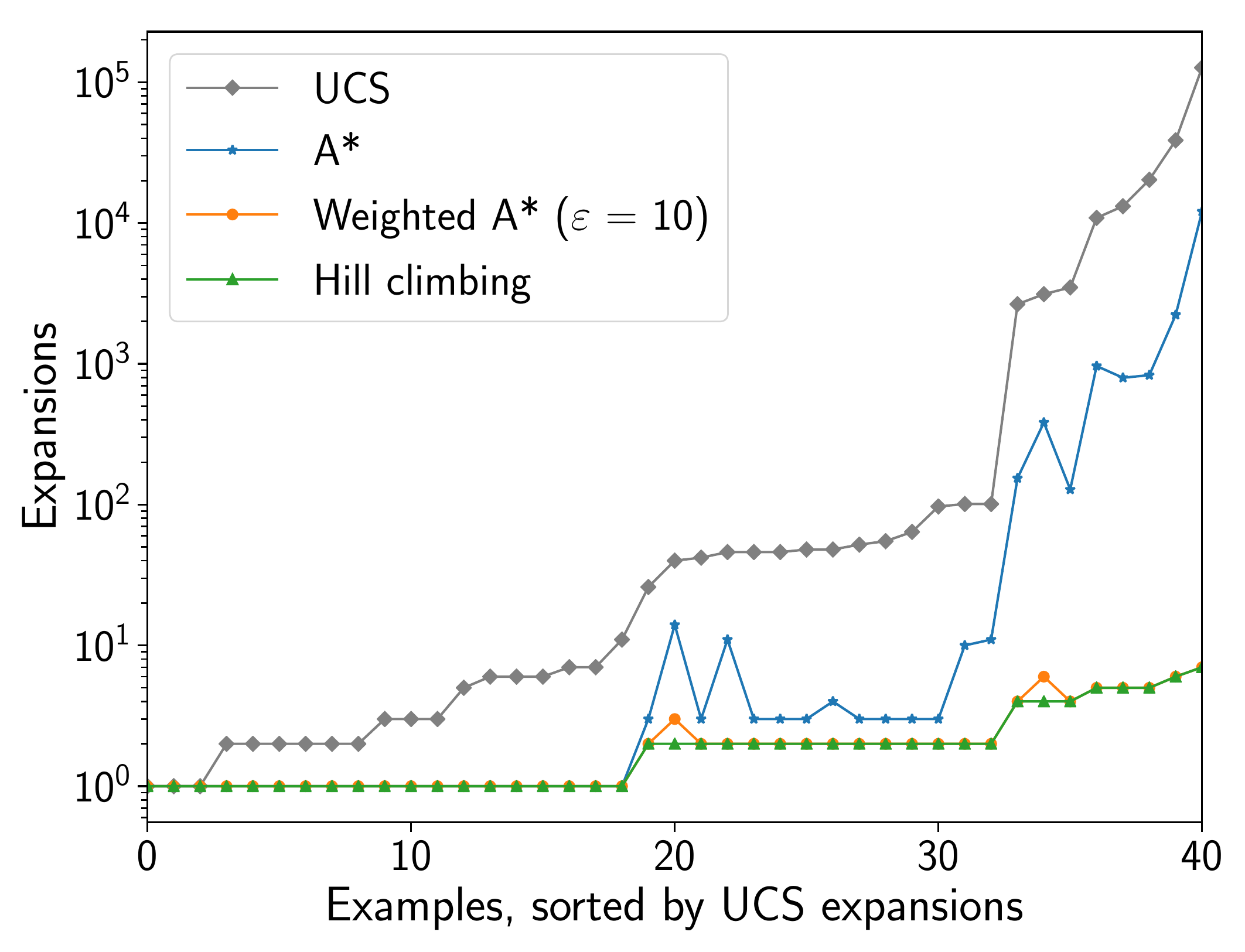}
        \quad
        \includegraphics[width=0.36\textwidth]{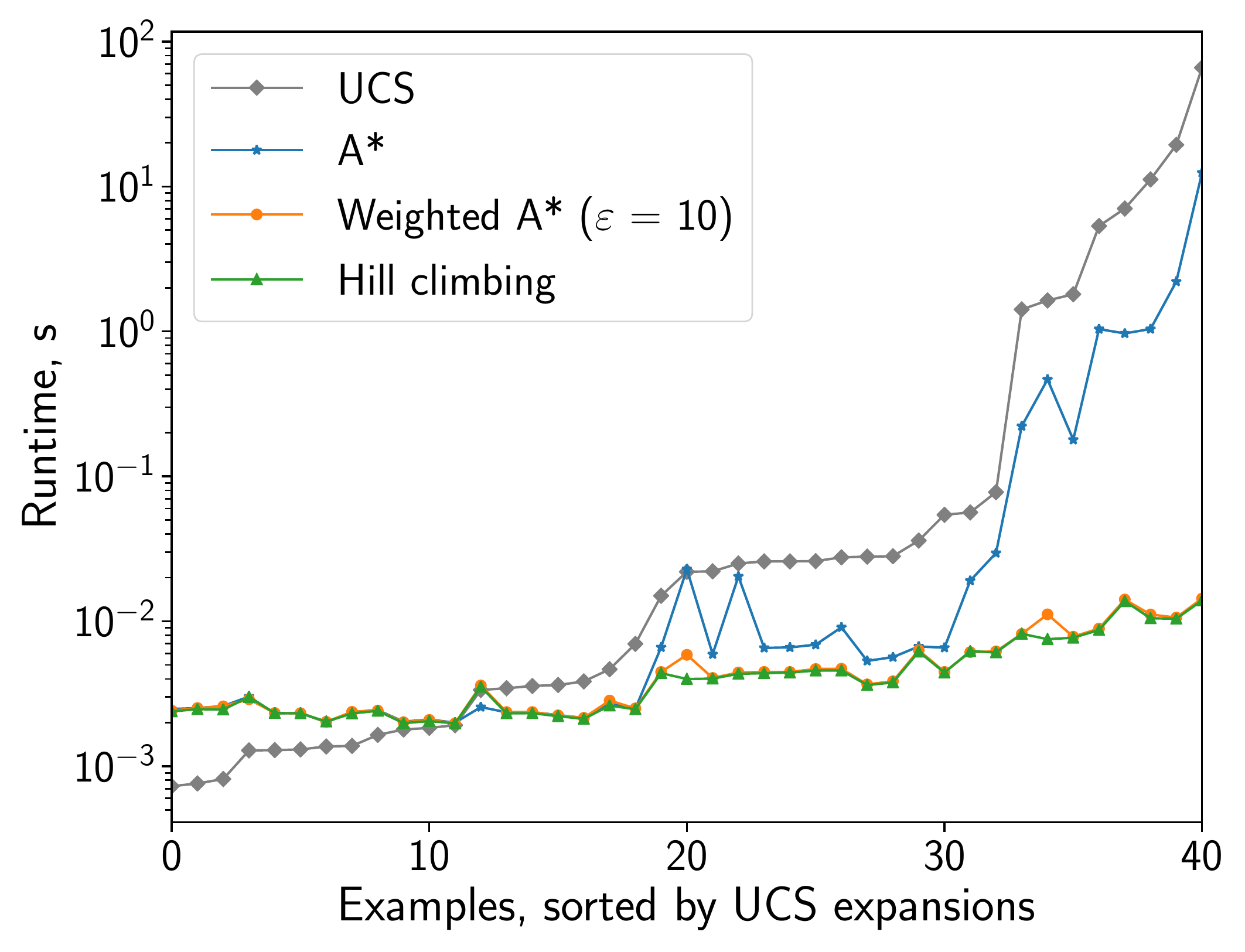}
    \caption{Basic attack setting comparison of UCS, \astar, weighted \astar, and hill climbing.
    Left: node expansions. Right: runtime in seconds. (y-axes are logarithmic)}
    \label{fig:bots-hill-climbing-perf}
\end{figure*}

We see in \Figref{fig:bots-hill-climbing-perf} that hill climbing performs significantly better
than UCS and \astar. Furthermore, we find that, in the basic-attack setting, all
adversarial examples found by the hill climbing incur minimal cost; and in the high-confidence
setting only some are more expensive (at most $1.2\times$ higher than the minimal cost).
We note that these results are on par with weighted \astar for $\varepsilon=10$, with the
difference that hill climbing does not provide any provable guarantees.

% \begin{figure}
%     \centering
%     \includegraphics[width=0.365\textwidth]{images/bots__hill_climbing__overhead__bins_20__band_1k__target_75.pdf}
%     \caption{Increase of path costs of adversarial examples found using hill climbing over
%     minimal costs in the high-confidence attack setting.}
%     \label{fig:bots-hill-climbing-overhead}
% \end{figure}

\descr{Realistic Adversarial Costs} Previously, we ensured that the chosen edge weights allow to use
admissible heuristics in \astar, and assumed that the adversary can modify all features at the
same cost. However, the general graphical framework and, more importantly, the problems it
represents in practice, are not limited to these transformation costs or such a powerful
adversary.

Here, we show how the graphical approach can accommodate a more realistic scenario. We constrain the
transformation graph to modify only the features that can be changed with the help of online services.
As of this writing, there exist online services that charge approximately \$2 for ghost-writing a
tweet or a reply; and services that charge approximately \$0.025 for a retweet or a like of a given
tweet. Hence, we constrain the adversary to modify only the \emph{number of tweets}, the
\emph{number of replies}, the \emph{likes per tweet}, and the \emph{retweets per tweet} features.
Moreover, we constrain the adversary to only increase the value of any transformable feature (e.g.,
we assume the adversary can hire someone to write more tweets, but not to delete them).

We set the weights of the edges such that they correspond to the dollar costs of the atomic
transformations. This cost is estimated as follows. We compute the difference between the previous
value of the feature and the lowest endpoint of the bucket in which the new feature value ends up.
We then multiply this difference (the number of tweets, replies, incoming likes, or incoming
retweets that need to be created or added) by the respective price in the mentioned online services.
As a result, path costs in such a graph are lower bounds on the dollar cost the adversary has to pay
to perform a sequence of transformations. Hence, these costs can be used to make informed risk
analysis regarding the security of a model.

\Tabref{tab:dollar-cost-results} shows the results of running UCS to obtain MAC values for the basic
and high-confidence attacks on this transformation graph. Because of the restricted
transformations, we can find adversarial examples only for 70\% and 19\% of the initial examples,
for the basic and the high-confidence setting, respectively.  In particular, we observe that if an
initial example is classified as ``bot'' with high enough confidence (approximately 80\% for the
basic attack), it is unlikely that we can find a corresponding adversarial example.

% For example, one could say that the model is rather secure against high-confidence
% attacks, since the success rate of such attacks (on the test dataset) is 19\%, and an average
% evasion cost is about \$60.

Even though for simplicity we used UCS, the edge weights could be expressed as a
\newterm{weighted norm}, e.g., $\norm{A(\vx- \vx')}_1$ for some positive-definite \newterm{weight
matrix} $A$ encoding the costs of the transformations. This means that it is possible to derive an
admissible heuristic and employ \astar. We leave the derivation of such heuristic as an open line
of research.

\begin{table}
    \caption{Dollar cost of adversarial examples against bot detection. Columns: \emph{Attack}---attack setting:
    required confidence level for adversarial examples; \emph{Exists}---proportion of initial examples
    for which an adversarial example exists; \emph{Minimal adversarial cost}---minimum, average, and maximum values of
    minimal adversarial costs for the part of the test dataset for which
    adversarial examples exist.} \label{tab:dollar-cost-results}
    \centering
    \resizebox{0.8\columnwidth}{!}{
    \begin{tabular}{l|r|rrr}
    \toprule
        &                         & \multicolumn{3}{l}{\textbf{Minimal adversarial cost}} \\
        \midrule
        \textbf{Attack} & \textbf{Exists} & min. & avg. & max.           \\
        \midrule
        Basic (50\%) & 70\%                    & \$0.02        & \$35.7        & \$281.6       \\
        High (75\%)  & 19\%                    & \$3.8         & \$57.6        & \$218.2       \\
        \bottomrule
    \end{tabular}
    }
\end{table}

% !TEX root = ../oakland/main.tex

\subsection{Building Defenses: Website Fingerprinting}
\label{sec:eval-wfp}

In the previous section, we considered a setting in which the adversary's knowledge is white-box.
This assumption, however, does not always hold in practice. When the access to the ML model is
black-box, the adversary cannot use the admissible heuristic to efficiently obtain minimal-cost
adversarial examples using \astar. We show that even in this setting the graphical framework is a
useful tool for finding adversarial examples in constrained domains.

In this section, we also change the adversarial perspective. We consider a scenario in which the ML
model is at the core of a privacy-invasive system and the entity deploying adversarial examples is
a target of this system. Therefore, the model becomes the adversary, and adversarial examples become
defenses.

Concretely, we take the case of \newterm{website fingerprinting} (WF), an attack in which a network
adversary attempts to infer which website a user is visiting by only looking at encrypted network
traffic~\cite{BackMS01, LiberatoreL06}, often using machine learning~\cite{PanchenkoNZE11,
WangCNJG14, HayesD16, SirinamIJW18, RimmerPJGJ18}.
This attack is mostly considered a threat to users of anonymous communication networks such as
Tor~\cite{DingledineMS04}. However, as the encrypted SNI proposal~\cite{ietf-tls-esni} becomes 
standardized within TLS~1.3~\cite{RFC8446}---hiding the destination of encrypted HTTPS traffic from
network observers---this attack becomes a privacy threat to all Internet users.
To counter the attack, existing ad-hoc defenses transform the traffic to reduce the accuracy of the
ML classifier~\cite{PanchenkoNZE11, DyerCRS12, CaiNWJG14, CaiNJ14, JuarezIPDW17}.

We first show how the graphical framework can be used as a systematic tool for designing traffic
modifications that defend users against a WF adversary. We then show how the
defenses produced by our method can be used as a baseline to evaluate both the effectiveness
(ability to fool a classifier) and efficiency (incurred overhead) of existing defenses.

\subsubsection{Website-Fingerprinting Attack}
We consider a WF adversary that takes as input a network \newterm{trace}, i.e., a sequence of
incoming (from server to client) and outgoing (from client to server) encrypted packets, and outputs
a binary guess of whether the user is visiting a website that is in the \newterm{monitored set} or
not. For instance, this could be a censorship adversary that wants to know if the visited website is
in a list of censored websites in order to stop the connection.

We simulate a WF adversary that uses the classifier by Panchenko et al.~\cite{PanchenkoNZE11}, an
SVM with an RBF kernel trained on \newterm{CUMUL features}. For a given trace, a CUMUL feature vector
contains the total incoming and outgoing packet counts, and 100 interpolated cumulative packet
counts. We refer the reader to the original paper for the details on computing the vector.

\descr{Dataset} We use the dataset of Tor network traces collected by \citet{WangCNJG14}. This
dataset contains 18,004 traces, half of them coming from a simulated monitored set compiled from 90
websites censored in China, the UK, and Saudi Arabia, and half coming from a non-monitored set of
5,000 other popular websites. The average trace length is 2,155.

We randomly split the dataset into 90\% training and 10\% testing subsets of 15,397 and 1,711
traces, respectively. We use these splits to train and test the target WF classifier. To keep the
running time of our experiments reasonable, we find adversarial examples only for traces with less
than 2000 packets that are classified as being in the monitored set. There are 577 such traces, with
an average length of 1750 packets.

The CUMUL classifier performs remarkably well on our test dataset, with an accuracy of 97.8\% (the
random baseline is 50\%). Thus, we consider it is a good example to illustrate the potential of our
framework.

\subsubsection{Building Defenses}

\descr{The Defender's Goals}
The goal of the defender is to modify monitored traces such that they are misclassified as
non-monitored by the WF classifier. These modifications can be of two kinds: adding dummy packets
and adding delay. Removing packets is not possible without affecting the content of the page, and
the delay is dependent on the network and cannot be decreased from an endpoint. The previous
work (e.g., \citet{PanchenkoNZE11}) has noted that perturbing timing information is not as important
to classification as the volume of packets. Hence, even though adding delay is possible within the
graphical framework, we consider that our defense \emph{only adds dummy packets}. As in the previous
works, we assume that dummy packets are filtered by the client and the server and do not affect the
application layer.

Adding dummy packets has a cost in terms of bandwidth and delay (routers have to process more
packets). Therefore, the defender wants to introduce as few packets as possible. In terms of confidence level
$\conflevel$, we consider only the case where the defender wants to flip the classifier decision
($\conflevel = \probthresh = 0.5$). Finding higher confidence adversarial examples is possible at
the cost of running a longer search.

\descr{The Defender's Capabilities: Transformation Graph and Cost}
For a given trace we define the following transformations: add one dummy outgoing packet, or one
dummy incoming packet, between any two existing packets in the trace. This means that each node in the
transformation graph is a copy of its parent trace with an added dummy packet. We assign every
transformation a cost equal to one, thus representing the added packet. Path
costs in this graph are equal to the number of added packets.

\descr{Heuristic and Search Algorithms}
Recall that CUMUL feature processing includes an interpolation step. Because of the interpolation
part of a CUMUL feature vector, the costs in our transformation graph can not be trivially expressed
as norm-induced distances between the feature vectors. Hence, we cannot use the admissible heuristic
from \Eqref{eq:heuristic-threshold}, or its approximated version. Instead, we use the following
\newterm{confidence-based heuristic}, similar to the heuristics used in other attacks in discrete
domains (e.g., by \citet{GaoLSQ18}):
\[h_t(x) = \begin{cases}
    +f(x),   & t = 0 \\
    -f(x),   & t = 1 \\
    -\infty, & F(x) = t \\
\end{cases}
\]
The value of this heuristic becomes lower as the confidence of the WF classifier for the target class
(non-monitored in our case) becomes higher. Note that this heuristic \emph{does not} require any
knowledge of the classifier, only the ability to query $f(x)$.

As in this case we cannot use optimal algorithms, we implement the hill climbing variation of the
greedy best-first search ($\mathsf{score}(x) = h(x)$ for its speed, see \Secref{sec:bfs}). We also
limit the number of iterations of the algorithm to 5,000 in order to keep down the runtime of
our experiments. Our results show that this is sufficient to find adversarial examples in 100\% of
the cases.

We also run a random search, i.e., we follow a random path in the graph until an adversarial example
is found, to obtain a baseline in terms of cost and runtime. We run this algorithm three times for
each trace with different random seeds.

\descr{Results} Hill climbing with the confidence-based heuristic finds adversarial traces in 100\%
of the cases, with an average time to find an adversarial example of 0.8 seconds. Random search
succeeds in slightly less than 100\%, with an average time of about 0.3 seconds. As we discuss below, the
results differ in the overhead required to find an adversarial example.

% We measure the performance in terms of the success rates---whether an
% adversarial example was found within the 10,000 iterations or not. Surprisingly, unlike in Twitter
% bots experiments where hill climbing has good performance (\Secref{sec:eval-bots}), for this problem
% and transformation graph the ability to backtrack is crucial.

\subsubsection{Comparing Defenses}

\begin{figure*}[t]
    \centering
    \includegraphics[width=0.402\textwidth]{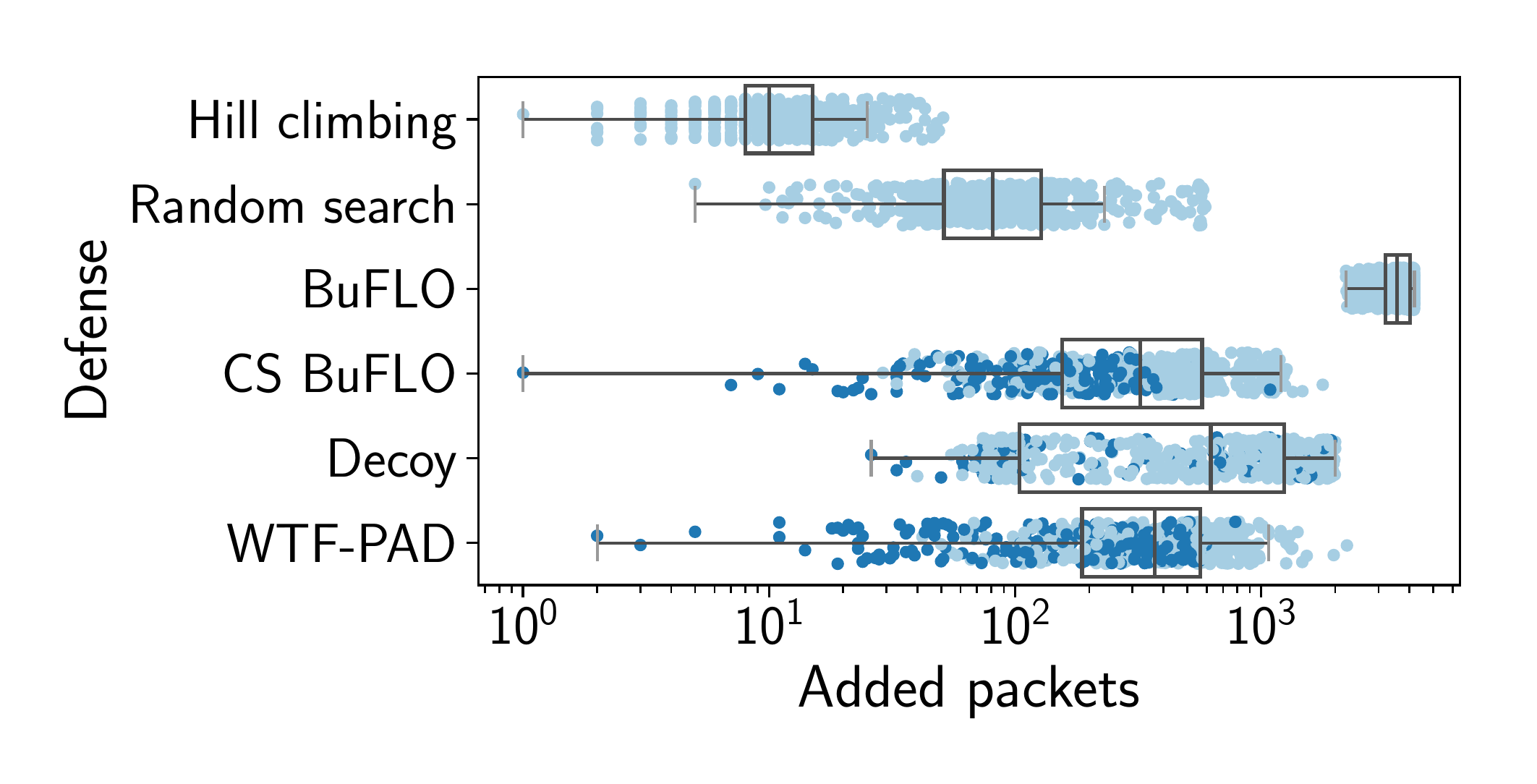}
    \includegraphics[width=0.28\textwidth,trim={6.5cm 0 0 0},clip]{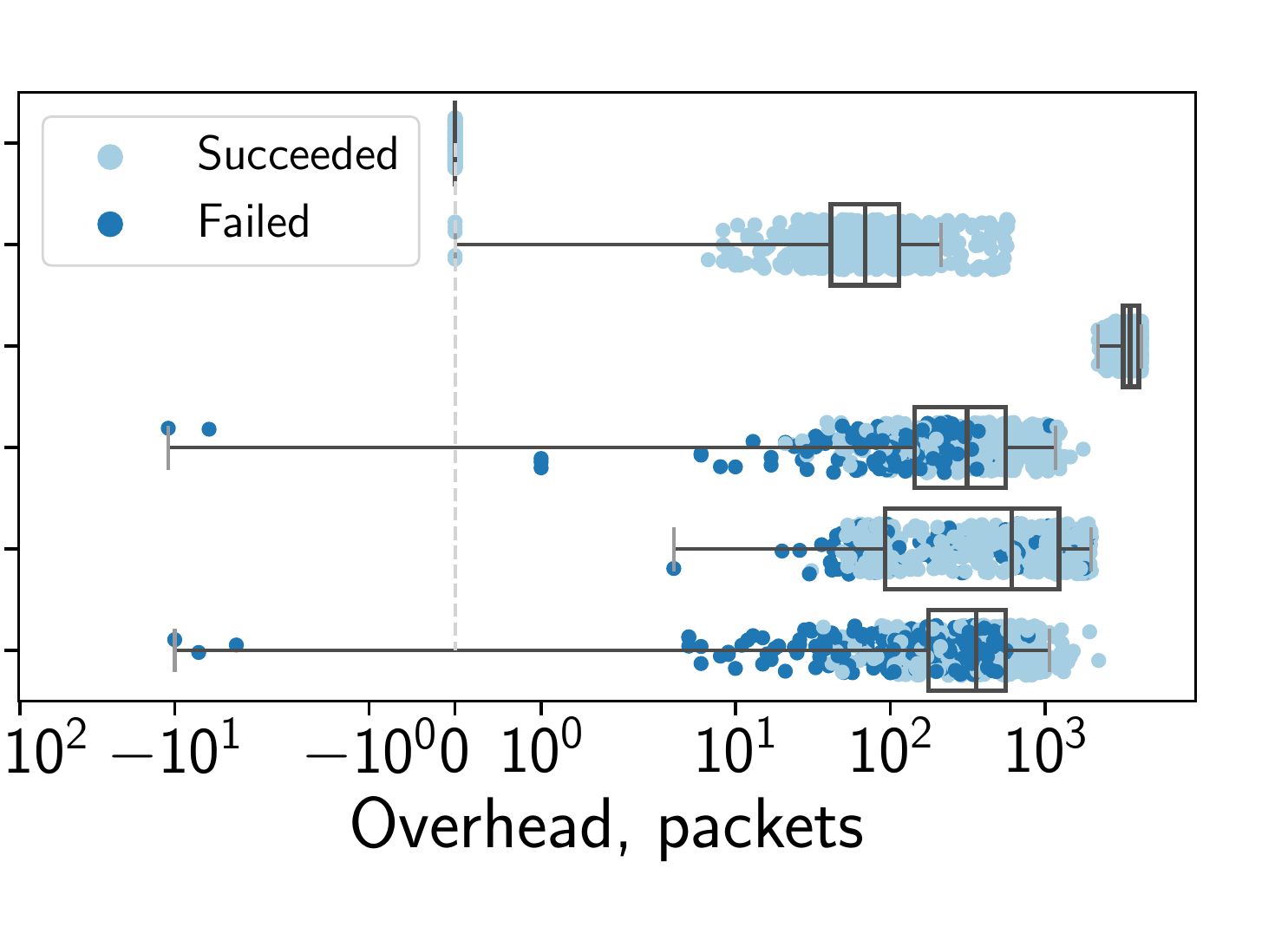}
    \includegraphics[width=0.28\textwidth,trim={6.5cm 0 0 0},clip]{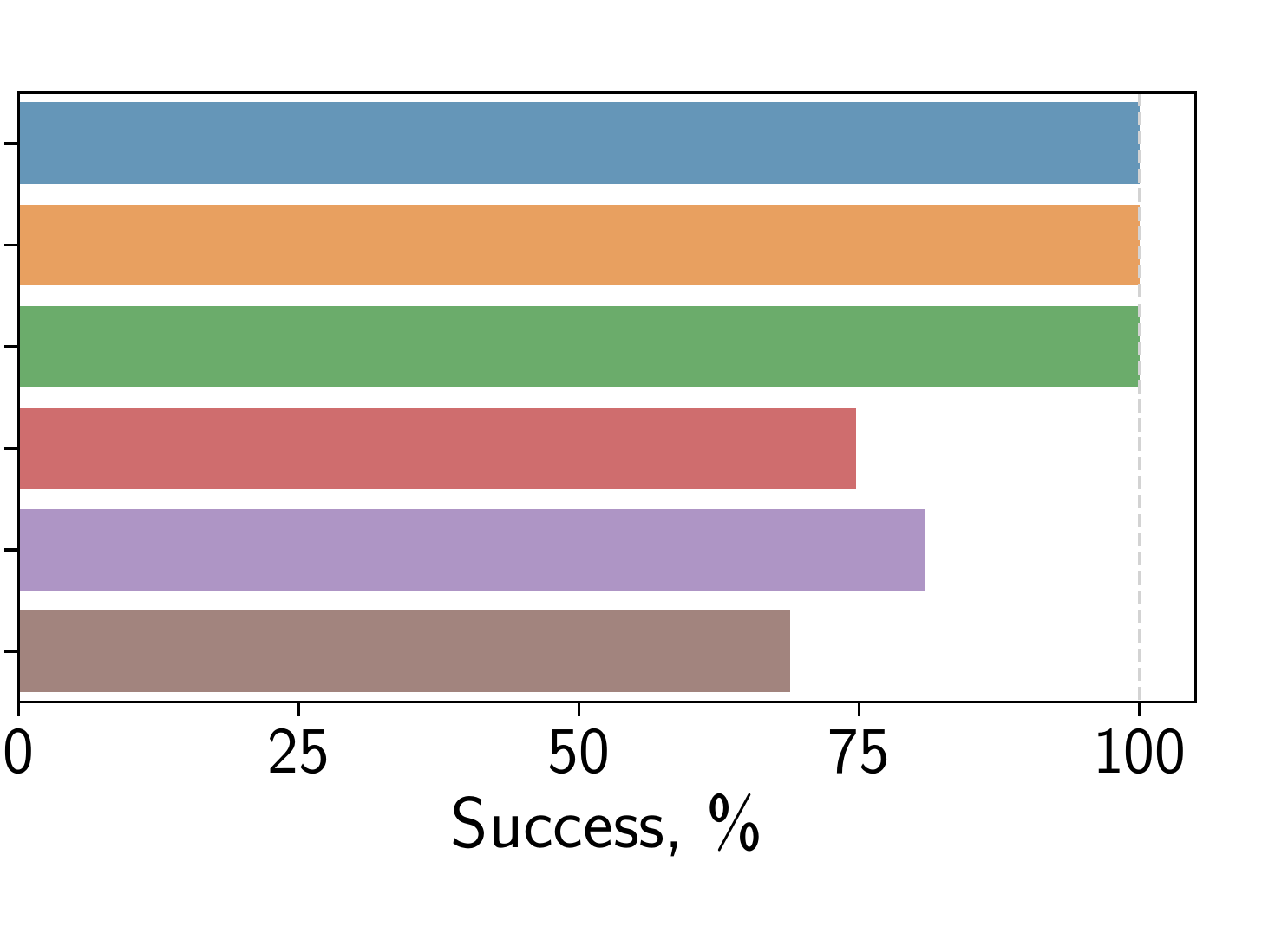}
    \caption{Left: number of added packets by WF defenses and adversarial examples (x-axis is
        logarithmic). Center: overhead of WF defenses compared to the number of packets added by
        adversarial examples found with hill climbing (x-axis is bi-symmetrically logarithmic).
        Right: success rates of WF defenses and adversarial examples against the SVM-RBF classifier.}
    \label{fig:wfp-defenses}
\end{figure*}

Here, we use minimal-cost adversarial examples for website fingerprinting to evaluate existing
ad-hoc WF defenses: Decoy pages~\cite{PanchenkoNZE11}, BuFLO~\cite{DyerCRS12}, CS
BuFLO~\cite{CaiNJ14}, and adaptive padding (WTF-PAD)~\cite{JuarezIPDW17}. We use the
implementations of Decoy pages and BuFLO by Wang,\footnote{\url{http://home.cse.ust.hk/~taow/wf/}}
CS BuFLO by \citet{Cherubin17},\footnote{\url{https://github.com/gchers/wfes}} and WTF-PAD by
\citet{JuarezIPDW17}.\footnote{\url{https://github.com/wtfpad/wtfpad}} Concretely, we measure the
overhead in terms of (1) the number of dummy packets added, and (2) the success rates of the
defenses, i.e., the percentage of traces for which they successfully evade the classifier.

We evaluate the efficiency by measuring the raw overhead in terms of added packets (see
\Figref{fig:wfp-defenses}, left).  The existing defenses add up to 3000 dummy packets.
Unexpectedly, BuFLO, which is deliberately inefficient, is the most expensive defense.  On the
contrary, adversarial examples add on average 12 dummy packets, and at most 52.

In terms of relative overhead, i.e., how many more packets the defenses add, compared to the
adversarial examples found using hill climbing, all defenses and random search require significantly
more bandwidth (see \Figref{fig:wfp-defenses}, center).  In five cases, CS BuFLO and WTF-PAD add
fewer packets, but in those cases the defenses do not succeed in evading the classifier. 

We then analyze the defenses' success rates in the light of the overhead they impose. The graph
search yields a 100\% success rate, whereas the existing defenses (aside from BuFLO) only succeed in
70\%---80\% of the cases (\Figref{fig:wfp-defenses}, right). Also, as it can be seen in
\Figref{fig:wfp-defenses} (center), increasing the number of packets is not a guarantee of success.
We see how for all defenses but BuFLO some cases fail even with hundreds of overhead packets. A
closer analysis shows that CS BuFLO and WTF-PAD often fail to defend shorter traces (under 1000
packets), whereas they can successfully defend the longer ones (see \Figref{fig:wfp-defenses-delta}
in \Appref{app:figures} for illustration). 

This hints that the ad-hoc defenses use the dummies in an inefficient way, and there is a
significant room for improvement. The adversarial examples found with hill climbing present a tight
upper bound on the minimal cost of any successful defense. Hence, we hope that our
graphical framework can serve as a baseline to evaluate the efficiency of future defenses, and guide
the design of effective website-fingerprinting countermeasures.

\subsubsection{Applicability Discussion}
In this section, we apply the graphical framework to a setting with no white-box knowledge, against
a non-linear classifier, by using non-optimal algorithms, and show that the framework is still
useful when none of the assumptions from \Secref{sec:eval-bots} hold.

We do not evaluate the transferability of obtained adversarial examples to other classifiers. The
main reason is that the state-of-the-art WF classifiers are based on deep learning \cite{RimmerPJGJ18,
SirinamIJW18}, thus require datasets larger than the one we use in our comparison. Although we leave
the transferability evaluation for future work, we expect that the results would be qualitatively
similar to those in the Twitter-bot case (see \Secref{sec:bots-discussion}).

% !TEX root = ../oakland/main.tex

\section{Related Work}
\label{sec:related}
We overview existing attacks in discrete domains and highlight their differences
with respect to our work.

\descr{Discretized Image Domain}
\citeauthor{PapernotMJFCS16} propose the Jacobian saliency map approach
(JSMA)~\cite{PapernotMJFCS16} to find adversarial images. JSMA is a greedy white-box attack that
transforms images by increasing or decreasing pixel intensity to maximize \newterm{saliency}, that 
is computed using the forward gradient of the target model. This attack is a basis for attacks in
other discrete domains~\cite{GrossePM0M16,JiaG18}, as we discuss below.

\descr{Text Domain} Multiple works study evasion attacks against text
classifiers~\cite{DalviDMSV04,LowdM05,PapernotMSH16, LiangSBLS18, HosseiniKZP17, GaoLSQ18,
EbrahimiRLD18, AlzantotSEHSC18}.  Recent attacks can be divided into three groups: those employing a
hill-climbing algorithm over the set of possible transformations of an initial piece of
text~\cite{PapernotMSH16, LiangSBLS18, HosseiniKZP17, GaoLSQ18}; those that greedily optimize the
forward gradient-based heuristic but run beam search~\cite{EbrahimiRLD18}; and those that use an
evolutionary algorithm~\cite{AlzantotSEHSC18}.

\descr{Malware Domain} Several works explore evasion attacks for malware, either adapting
JSMA~\cite{GrossePM0M16}, applying forward gradient-based heuristics~\cite{KolosnjajiDBMGER18},
using a black-box hill-climbing algorithm over a set of feasible transformations~\cite{DangHC17}, or
using a black-box evolutionary algorithm~\cite{XuQE16}.

\descr{Protecting Users}
Finally, some works use adversarial examples as means to protect users.  \citeauthor{JiaG18}~adapt
JSMA to modify user-item relationship vectors in the context of recommendation systems.
\citeauthor{OverdorfKBTG18}~use exhaustive search to find adversarial examples that counter
anti-social effects of machine learning.

\subsection{Comparison to Our Work}
\label{sec:related-instantiations}

Our framework can be seen as a generalization of most of the attacks mentioned previously. Moreover,
it can encode arbitrary adversarial costs, and can be configured to output minimal-cost adversarial
examples using \astar search. We note that in parallel to our work, \citeauthor{WuWRHK18} also used
\astar and randomized tree search to obtain bounds on robustness of neural networks~\cite{WuWRHK18}.
Their work, however, only considers the setting of image recognition.

\citet{DalviDMSV04} have used integer linear programming to find approximate minimal-cost
adversarial examples against a Na\"ive Bayes classifier. Our graphical framework enables us to
consider arbitrary transformations and cost models and to find minimal-cost adversarial examples
for any non-linear classifier for which adversarial robustness in a continuous domain can be
computed.

Except for the method by \citeauthor{DalviDMSV04} and the attacks based on evolutionary algorithms,
all of the attacks above can be instantiated in our graphical framework (see \Tabref{tab:comparison}
in \Appref{app:figures} for a summary of such instantiations).  The attacks implicitly define a
transformation graph by specifying a set of domain-specific transformations (e.g., word insertions
for text) that define the graph. The cost of transformations can be equal to the number of such
transformations (e.g., \cite{PapernotMSH16, LiangSBLS18}), or, equivalently, to the $\lp$ distance
between feature vectors interpreted as the number of transformations (e.g., \cite{GrossePM0M16,
JiaG18}).

The attacks can be seen as special cases of running the BF$^*$ search algorithm (see
\Secref{sec:bfs}) over a transformation graph. They differ in adversarial knowledge assumptions
(white-box or black-box), transformation graphs, adversarial cost models, and the choice of the
scoring function and priority-queue capacity that defines the instantiation of BF$^*$.

Most of the attacks~(e.g., \cite{PapernotMSH16, LiangSBLS18, GrossePM0M16, JiaG18}) run a
hill-climbing search over the transformation graph. They maximize a heuristic either based
on the forward gradient of the model (in the white-box setting where the adversary can compute the
gradient), or on the confidence (in the black-box setting where the adversary can only query the
model). \citet{EbrahimiRLD18} use beam search instead of hill climbing, \citet{textfool} uses an
instance of backtracking best-first search, and \citet{OverdorfKBTG18} use an exhaustive search over
the space of feasible transformations, equivalent to UCS.

% !TEX root = ../oakland/main.tex
\section{Conclusions}
In this paper, we proposed a graphical framework for formalizing evasion attacks in discrete
domains.  This framework casts attacks as search over a graph of valid transformations of an initial
example. It generalizes many proposed attacks in various discrete domains, and offers additional
benefits.

First, as a formalization, it enables us to define arbitrary adversarial costs and to choose search
algorithms from the vast literature on graph search, whereas the previous attacks often use $\lp$
norms as costs and mostly focus on hill-climbing strategies.

Second, we show that when it is possible to compute adversarial robustness in a continuous domain,
this robustness measure can be used as a heuristic to efficiently explore a discrete domain.
Thus, an adversary with the white-box knowledge can use \astar search to obtain adversarial examples
that incur minimal adversarial cost. This enables us to provably evaluate the adversarial robustness
of a classifier given the domain constraints and adversary's capabilities. 

Third, the versatility of our framework to model transformations and their costs independently of the
ML model under attack make it suitable to tackle both security and privacy problems. As examples,
we showed how it can be used to evaluate the adversarial robustness of a Twitter-bot classifier, and to
evaluate the cost-effectiveness of privacy defenses against a website-fingerprinting classifier.

\vspace{2em}

\section*{Acknowledgements}
We would like to thank Danesh Irani, \'Ulfar Erlingsson, Alexey Kurakin, Seyed-Mohsen
Moosavi-Dezfooli, Maksym Andriushchenko, and Giovanni Cherubin for their feedback and helpful discussions. This
research is funded by NEXTLEAP project\footnote{\url{https://nextleap.eu}} within the European
Union's Horizon 2020 Framework Program for Research and Innovation (H2020-ICT-2015, ICT-10-2015)
under grant agreement 688722. Jamie Hayes is funded by a Google PhD Fellowship in Machine Learning.

{\footnotesize
\bibliographystyle{IEEEtranN}
\bibliography{include/main}

% Generated by IEEEtranN.bst, version: 1.14 (2015/08/26)
\begin{thebibliography}{71}
\providecommand{\natexlab}[1]{#1}
\providecommand{\url}[1]{#1}
\csname url@samestyle\endcsname
\providecommand{\newblock}{\relax}
\providecommand{\bibinfo}[2]{#2}
\providecommand{\BIBentrySTDinterwordspacing}{\spaceskip=0pt\relax}
\providecommand{\BIBentryALTinterwordstretchfactor}{4}
\providecommand{\BIBentryALTinterwordspacing}{\spaceskip=\fontdimen2\font plus
\BIBentryALTinterwordstretchfactor\fontdimen3\font minus
  \fontdimen4\font\relax}
\providecommand{\BIBforeignlanguage}[2]{{%
\expandafter\ifx\csname l@#1\endcsname\relax
\typeout{** WARNING: IEEEtranN.bst: No hyphenation pattern has been}%
\typeout{** loaded for the language `#1'. Using the pattern for}%
\typeout{** the default language instead.}%
\else
\language=\csname l@#1\endcsname
\fi
#2}}
\providecommand{\BIBdecl}{\relax}
\BIBdecl

\bibitem[Madry et~al.(2017)Madry, Makelov, Schmidt, Tsipras, and
  Vladu]{MadryMSTV17}
\BIBentryALTinterwordspacing
A.~Madry, A.~Makelov, L.~Schmidt, D.~Tsipras, and A.~Vladu, ``Towards deep
  learning models resistant to adversarial attacks,'' \emph{CoRR}, vol.
  abs/1706.06083, 2017.
\BIBentrySTDinterwordspacing

\bibitem[Carlini and Wagner(2017)]{CarliniWagner17}
\BIBentryALTinterwordspacing
N.~Carlini and D.~A. Wagner, ``Towards evaluating the robustness of neural
  networks,'' in \emph{2017 {IEEE} Symposium on Security and Privacy, {SP}
  2017, San Jose, CA, USA, May 22-26, 2017}.\hskip 1em plus 0.5em minus
  0.4em\relax {IEEE} Computer Society, 2017, 2017.
\BIBentrySTDinterwordspacing

\bibitem[Moosavi{-}Dezfooli et~al.(2016)Moosavi{-}Dezfooli, Fawzi, and
  Frossard]{Moosavi-Dezfooli16}
\BIBentryALTinterwordspacing
S.~Moosavi{-}Dezfooli, A.~Fawzi, and P.~Frossard, ``Deepfool: {A} simple and
  accurate method to fool deep neural networks,'' in \emph{2016 {IEEE}
  Conference on Computer Vision and Pattern Recognition, {CVPR} 2016, Las
  Vegas, NV, USA, June 27-30, 2016}.\hskip 1em plus 0.5em minus 0.4em\relax
  {IEEE} Computer Society, 2016, 2016.
\BIBentrySTDinterwordspacing

\bibitem[Papernot et~al.(2016{\natexlab{a}})Papernot, McDaniel, Jha,
  Fredrikson, Celik, and Swami]{PapernotMJFCS16}
\BIBentryALTinterwordspacing
N.~Papernot, P.~D. McDaniel, S.~Jha, M.~Fredrikson, Z.~B. Celik, and A.~Swami,
  ``The limitations of deep learning in adversarial settings,'' in \emph{{IEEE}
  European Symposium on Security and Privacy, EuroS{\&}P 2016,
  Saarbr{\"{u}}cken, Germany, March 21-24, 2016}.\hskip 1em plus 0.5em minus
  0.4em\relax {IEEE}, 2016, 2016.
\BIBentrySTDinterwordspacing

\bibitem[Goodfellow et~al.(2014)Goodfellow, Shlens, and
  Szegedy]{GoodfellowSS14}
\BIBentryALTinterwordspacing
I.~J. Goodfellow, J.~Shlens, and C.~Szegedy, ``Explaining and harnessing
  adversarial examples,'' \emph{CoRR}, vol. abs/1412.6572, 2014.
\BIBentrySTDinterwordspacing

\bibitem[Biggio et~al.(2013)Biggio, Corona, Maiorca, Nelson, Srndic, Laskov,
  Giacinto, and Roli]{BiggioCMNSLGR13}
\BIBentryALTinterwordspacing
B.~Biggio, I.~Corona, D.~Maiorca, B.~Nelson, N.~Srndic, P.~Laskov, G.~Giacinto,
  and F.~Roli, ``Evasion attacks against machine learning at test time,'' in
  \emph{Machine Learning and Knowledge Discovery in Databases - European
  Conference, {ECML} {PKDD} 2013, Prague, Czech Republic, September 23-27,
  2013, Proceedings, Part {III}}, ser. Lecture Notes in Computer Science, vol.
  8190.\hskip 1em plus 0.5em minus 0.4em\relax Springer, 2013, 2013.
\BIBentrySTDinterwordspacing

\bibitem[Szegedy et~al.(2013)Szegedy, Zaremba, Sutskever, Bruna, Erhan,
  Goodfellow, and Fergus]{SzegedyZSBEGF13}
\BIBentryALTinterwordspacing
C.~Szegedy, W.~Zaremba, I.~Sutskever, J.~Bruna, D.~Erhan, I.~J. Goodfellow, and
  R.~Fergus, ``Intriguing properties of neural networks,'' \emph{CoRR}, vol.
  abs/1312.6199, 2013.
\BIBentrySTDinterwordspacing

\bibitem[Bastani et~al.(2016)Bastani, Ioannou, Lampropoulos, Vytiniotis, Nori,
  and Criminisi]{BastaniILVNC16}
\BIBentryALTinterwordspacing
O.~Bastani, Y.~Ioannou, L.~Lampropoulos, D.~Vytiniotis, A.~V. Nori, and
  A.~Criminisi, ``Measuring neural net robustness with constraints,'' in
  \emph{Advances in Neural Information Processing Systems 29: Annual Conference
  on Neural Information Processing Systems 2016, December 5-10, 2016,
  Barcelona, Spain}, 2016, 2016.
\BIBentrySTDinterwordspacing

\bibitem[Katz et~al.(2017)Katz, Barrett, Dill, Julian, and
  Kochenderfer]{KatzBDJK17}
\BIBentryALTinterwordspacing
G.~Katz, C.~W. Barrett, D.~L. Dill, K.~Julian, and M.~J. Kochenderfer,
  ``Reluplex: An efficient {SMT} solver for verifying deep neural networks,''
  in \emph{Computer Aided Verification - 29th International Conference, {CAV}
  2017, Heidelberg, Germany, July 24-28, 2017, Proceedings, Part {I}}, ser.
  Lecture Notes in Computer Science, vol. 10426.\hskip 1em plus 0.5em minus
  0.4em\relax Springer, 2017, 2017.
\BIBentrySTDinterwordspacing

\bibitem[Hein and Andriushchenko(2017)]{HeinA17}
\BIBentryALTinterwordspacing
M.~Hein and M.~Andriushchenko, ``Formal guarantees on the robustness of a
  classifier against adversarial manipulation,'' in \emph{Advances in Neural
  Information Processing Systems 30: Annual Conference on Neural Information
  Processing Systems 2017, 4-9 December 2017, Long Beach, CA, {USA}}, 2017,
  2017.
\BIBentrySTDinterwordspacing

\bibitem[Fawzi et~al.(2018)Fawzi, Fawzi, and Frossard]{FawziFF18}
\BIBentryALTinterwordspacing
A.~Fawzi, O.~Fawzi, and P.~Frossard, ``Analysis of classifiers' robustness to
  adversarial perturbations,'' \emph{Machine Learning}, vol. 107, no.~3, pp.
  481--508, 2018.
\BIBentrySTDinterwordspacing

\bibitem[Wong et~al.(2018)Wong, Schmidt, Metzen, and Kolter]{WongSMK18}
\BIBentryALTinterwordspacing
E.~Wong, F.~Schmidt, J.~H. Metzen, and J.~Z. Kolter, ``Scaling provable
  adversarial defenses,'' \emph{CoRR}, vol. abs/1805.12514, 2018.
\BIBentrySTDinterwordspacing

\bibitem[Tsuzuku et~al.(2018)Tsuzuku, Sato, and Sugiyama]{TsuzukuSS18}
\BIBentryALTinterwordspacing
Y.~Tsuzuku, I.~Sato, and M.~Sugiyama, ``Lipschitz-margin training: Scalable
  certification of perturbation invariance for deep neural networks,'' in
  \emph{Advances in Neural Information Processing Systems 31: Annual Conference
  on Neural Information Processing Systems 2018, NeurIPS 2018, 3-8 December
  2018, Montr{\'{e}}al, Canada.}, 2018, 2018.
\BIBentrySTDinterwordspacing

\bibitem[Demontis et~al.(2017)Demontis, Melis, Biggio, Maiorca, Arp, Rieck,
  Corona, Giacinto, and Roli]{DemontisMBMARCG17}
\BIBentryALTinterwordspacing
A.~Demontis, M.~Melis, B.~Biggio, D.~Maiorca, D.~Arp, K.~Rieck, I.~Corona,
  G.~Giacinto, and F.~Roli, ``Yes, machine learning can be more secure! {A}
  case study on android malware detection,'' \emph{CoRR}, vol. abs/1704.08996,
  2017.
\BIBentrySTDinterwordspacing

\bibitem[Ebrahimi et~al.(2018)Ebrahimi, Rao, Lowd, and Dou]{EbrahimiRLD18}
\BIBentryALTinterwordspacing
J.~Ebrahimi, A.~Rao, D.~Lowd, and D.~Dou, ``Hotflip: White-box adversarial
  examples for text classification,'' in \emph{Proceedings of the 56th Annual
  Meeting of the Association for Computational Linguistics, {ACL} 2018,
  Melbourne, Australia, July 15-20, 2018, Volume 2: Short Papers}.\hskip 1em
  plus 0.5em minus 0.4em\relax Association for Computational Linguistics, 2018,
  2018.
\BIBentrySTDinterwordspacing

\bibitem[Jia and Gong(2018)]{JiaG18}
\BIBentryALTinterwordspacing
J.~Jia and N.~Z. Gong, ``Attriguard: {A} practical defense against attribute
  inference attacks via adversarial machine learning,'' in \emph{27th {USENIX}
  Security Symposium, {USENIX} Security 2018, Baltimore, MD, USA, August 15-17,
  2018.}\hskip 1em plus 0.5em minus 0.4em\relax {USENIX} Association, 2018,
  2018.
\BIBentrySTDinterwordspacing

\bibitem[Kolosnjaji et~al.(2018)Kolosnjaji, Demontis, Biggio, Maiorca,
  Giacinto, Eckert, and Roli]{KolosnjajiDBMGER18}
\BIBentryALTinterwordspacing
B.~Kolosnjaji, A.~Demontis, B.~Biggio, D.~Maiorca, G.~Giacinto, C.~Eckert, and
  F.~Roli, ``Adversarial malware binaries: Evading deep learning for malware
  detection in executables,'' \emph{CoRR}, vol. abs/1803.04173, 2018.
\BIBentrySTDinterwordspacing

\bibitem[Miyato et~al.(2016)Miyato, Dai, and Goodfellow]{MiyatoDG16}
\BIBentryALTinterwordspacing
T.~Miyato, A.~M. Dai, and I.~J. Goodfellow, ``Virtual adversarial training for
  semi-supervised text classification,'' \emph{CoRR}, vol. abs/1605.07725,
  2016.
\BIBentrySTDinterwordspacing

\bibitem[Papernot et~al.(2016{\natexlab{b}})Papernot, McDaniel, Swami, and
  Harang]{PapernotMSH16}
\BIBentryALTinterwordspacing
N.~Papernot, P.~D. McDaniel, A.~Swami, and R.~E. Harang, ``Crafting adversarial
  input sequences for recurrent neural networks,'' in \emph{2016 {IEEE}
  Military Communications Conference, {MILCOM} 2016, Baltimore, MD, USA,
  November 1-3, 2016}.\hskip 1em plus 0.5em minus 0.4em\relax {IEEE}, 2016,
  2016.
\BIBentrySTDinterwordspacing

\bibitem[Grosse et~al.(2016)Grosse, Papernot, Manoharan, Backes, and
  McDaniel]{GrossePM0M16}
\BIBentryALTinterwordspacing
K.~Grosse, N.~Papernot, P.~Manoharan, M.~Backes, and P.~D. McDaniel,
  ``Adversarial perturbations against deep neural networks for malware
  classification,'' \emph{CoRR}, vol. abs/1606.04435, 2016.
\BIBentrySTDinterwordspacing

\bibitem[Liang et~al.(2018)Liang, Li, Su, Bian, Li, and Shi]{LiangSBLS18}
\BIBentryALTinterwordspacing
B.~Liang, H.~Li, M.~Su, P.~Bian, X.~Li, and W.~Shi, ``Deep text classification
  can be fooled,'' in \emph{Proceedings of the Twenty-Seventh International
  Joint Conference on Artificial Intelligence, {IJCAI} 2018, July 13-19, 2018,
  Stockholm, Sweden.}\hskip 1em plus 0.5em minus 0.4em\relax ijcai.org, 2018,
  2018.
\BIBentrySTDinterwordspacing

\bibitem[Gao et~al.(2018)Gao, Lanchantin, Soffa, and Qi]{GaoLSQ18}
\BIBentryALTinterwordspacing
J.~Gao, J.~Lanchantin, M.~L. Soffa, and Y.~Qi, ``Black-box generation of
  adversarial text sequences to evade deep learning classifiers,'' in
  \emph{2018 {IEEE} Security and Privacy Workshops, {SP} Workshops 2018, San
  Francisco, CA, USA, May 24, 2018}.\hskip 1em plus 0.5em minus 0.4em\relax
  {IEEE}, 2018, 2018.
\BIBentrySTDinterwordspacing

\bibitem[Overdorf et~al.(2018)Overdorf, Kulynych, Balsa, Troncoso, and
  G{\"u}rses]{OverdorfKBTG18}
R.~Overdorf, B.~Kulynych, E.~Balsa, C.~Troncoso, and S.~G{\"u}rses, ``{POTs}:
  Protective optimization technologies,'' \emph{CoRR}, vol. abs/1806.02711,
  2018.

\bibitem[Lowd and Meek(2005)]{LowdM05}
D.~Lowd and C.~Meek, ``Adversarial learning,'' in \emph{{SIGKDD}}, 2005, 2005.

\bibitem[Asif et~al.(2015)Asif, Xing, Behpour, and Ziebart]{AsifXBZ15}
\BIBentryALTinterwordspacing
K.~Asif, W.~Xing, S.~Behpour, and B.~D. Ziebart, ``Adversarial cost-sensitive
  classification,'' in \emph{Proceedings of the Thirty-First Conference on
  Uncertainty in Artificial Intelligence, {UAI} 2015, July 12-16, 2015,
  Amsterdam, The Netherlands}.\hskip 1em plus 0.5em minus 0.4em\relax {AUAI}
  Press, 2015, 2015.
\BIBentrySTDinterwordspacing

\bibitem[Zhang and Evans(2018)]{ZhangE19}
\BIBentryALTinterwordspacing
X.~Zhang and D.~Evans, ``Cost-sensitive robustness against adversarial
  examples,'' \emph{CoRR}, vol. abs/1810.09225, 2018.
\BIBentrySTDinterwordspacing

\bibitem[Cadwalladr and Graham-Harrison(2018)]{Cadwalladr18}
C.~Cadwalladr and E.~Graham-Harrison, ``How {Cambridge Analytica} turned
  facebook ‘likes’ into a lucrative political tool,'' \emph{Guardian},
  2018.

\bibitem[Abbasi and Chen(2008)]{AbbasiC08}
\BIBentryALTinterwordspacing
A.~Abbasi and H.~Chen, ``Writeprints: {A} stylometric approach to
  identity-level identification and similarity detection in cyberspace,''
  \emph{{ACM} Trans. Inf. Syst.}, vol.~26, no.~2, pp. 7:1--7:29, 2008.
\BIBentrySTDinterwordspacing

\bibitem[Kosinski et~al.(2013)Kosinski, Stillwell, and Graepel]{KosinskiSG13}
M.~Kosinski, D.~Stillwell, and T.~Graepel, ``Private traits and attributes are
  predictable from digital records of human behavior,'' \emph{Proceedings of
  the National Academy of Sciences}, p. 201218772, 2013.

\bibitem[HRW(2018)]{HRW18}
``Coalition letter to {Amazon} regarding the facial recognition system,
  {Rekognition},'' 2018.

\bibitem[Back et~al.(2001)Back, M{\"{o}}ller, and Stiglic]{BackMS01}
\BIBentryALTinterwordspacing
A.~Back, U.~M{\"{o}}ller, and A.~Stiglic, ``Traffic analysis attacks and
  trade-offs in anonymity providing systems,'' in \emph{Information Hiding, 4th
  International Workshop, {IHW} 2001, Pittsburgh, PA, USA, April 25-27, 2001,
  Proceedings}, ser. Lecture Notes in Computer Science, vol. 2137.\hskip 1em
  plus 0.5em minus 0.4em\relax Springer, 2001, 2001.
\BIBentrySTDinterwordspacing

\bibitem[Liberatore and Levine(2006)]{LiberatoreL06}
\BIBentryALTinterwordspacing
M.~Liberatore and B.~N. Levine, ``Inferring the source of encrypted {HTTP}
  connections,'' in \emph{Proceedings of the 13th {ACM} Conference on Computer
  and Communications Security, {CCS} 2006, Alexandria, VA, USA, October 30 -
  November 3, 2006}.\hskip 1em plus 0.5em minus 0.4em\relax {ACM}, 2006, 2006.
\BIBentrySTDinterwordspacing

\bibitem[Panchenko et~al.(2011)Panchenko, Niessen, Zinnen, and
  Engel]{PanchenkoNZE11}
\BIBentryALTinterwordspacing
A.~Panchenko, L.~Niessen, A.~Zinnen, and T.~Engel, ``Website fingerprinting in
  onion routing based anonymization networks,'' in \emph{Proceedings of the
  10th annual {ACM} workshop on Privacy in the electronic society, {WPES} 2011,
  Chicago, IL, USA, October 17, 2011}.\hskip 1em plus 0.5em minus 0.4em\relax
  {ACM}, 2011, 2011.
\BIBentrySTDinterwordspacing

\bibitem[Dyer et~al.(2012)Dyer, Coull, Ristenpart, and Shrimpton]{DyerCRS12}
\BIBentryALTinterwordspacing
K.~P. Dyer, S.~E. Coull, T.~Ristenpart, and T.~Shrimpton, ``Peek-a-boo, {I}
  still see you: Why efficient traffic analysis countermeasures fail,'' in
  \emph{{IEEE} Symposium on Security and Privacy, {SP} 2012, 21-23 May 2012,
  San Francisco, California, {USA}}.\hskip 1em plus 0.5em minus 0.4em\relax
  {IEEE} Computer Society, 2012, 2012.
\BIBentrySTDinterwordspacing

\bibitem[Cai et~al.(2014{\natexlab{a}})Cai, Nithyanand, Wang, Johnson, and
  Goldberg]{CaiNWJG14}
\BIBentryALTinterwordspacing
X.~Cai, R.~Nithyanand, T.~Wang, R.~Johnson, and I.~Goldberg, ``A systematic
  approach to developing and evaluating website fingerprinting defenses,'' in
  \emph{Proceedings of the 2014 {ACM} {SIGSAC} Conference on Computer and
  Communications Security, Scottsdale, AZ, USA, November 3-7, 2014}.\hskip 1em
  plus 0.5em minus 0.4em\relax {ACM}, 2014, 2014.
\BIBentrySTDinterwordspacing

\bibitem[Cai et~al.(2014{\natexlab{b}})Cai, Nithyanand, and Johnson]{CaiNJ14}
\BIBentryALTinterwordspacing
X.~Cai, R.~Nithyanand, and R.~Johnson, ``{CS-BuFLO}: {A} congestion sensitive
  website fingerprinting defense,'' in \emph{Proceedings of the 13th Workshop
  on Privacy in the Electronic Society, {WPES} 2014, Scottsdale, AZ, USA,
  November 3, 2014}.\hskip 1em plus 0.5em minus 0.4em\relax {ACM}, 2014, 2014.
\BIBentrySTDinterwordspacing

\bibitem[Ju{\'{a}}rez et~al.(2016)Ju{\'{a}}rez, Imani, Perry, D{\'{\i}}az, and
  Wright]{JuarezIPDW17}
\BIBentryALTinterwordspacing
M.~Ju{\'{a}}rez, M.~Imani, M.~Perry, C.~D{\'{\i}}az, and M.~Wright, ``Toward an
  efficient website fingerprinting defense,'' in \emph{Computer Security -
  {ESORICS} 2016 - 21st European Symposium on Research in Computer Security,
  Heraklion, Greece, September 26-30, 2016, Proceedings, Part {I}}, ser.
  Lecture Notes in Computer Science, vol. 9878.\hskip 1em plus 0.5em minus
  0.4em\relax Springer, 2016, 2016.
\BIBentrySTDinterwordspacing

\bibitem[Dechter and Pearl(1985)]{DechterP85}
\BIBentryALTinterwordspacing
R.~Dechter and J.~Pearl, ``Generalized best-first search strategies and the
  optimality of {A*},'' \emph{J. {ACM}}, vol.~32, no.~3, pp. 505--536, 1985.
\BIBentrySTDinterwordspacing

\bibitem[Hart et~al.(1968)Hart, Nilsson, and Raphael]{HartNR68}
\BIBentryALTinterwordspacing
P.~E. Hart, N.~J. Nilsson, and B.~Raphael, ``A formal basis for the heuristic
  determination of minimum cost paths,'' \emph{{IEEE} Trans. Systems Science
  and Cybernetics}, vol.~4, no.~2, pp. 100--107, 1968.
\BIBentrySTDinterwordspacing

\bibitem[Doran and Michie(1966)]{DoranMichie66}
J.~E. Doran and D.~Michie, ``Experiments with the graph traverser program,''
  \emph{Proc. R. Soc. Lond. A}, vol. 294, no. 1437, pp. 235--259, 1966.

\bibitem[Pohl(1970)]{Pohl70}
\BIBentryALTinterwordspacing
I.~Pohl, ``Heuristic search viewed as path finding in a graph,'' \emph{Artif.
  Intell.}, vol.~1, no.~3, pp. 193--204, 1970.
\BIBentrySTDinterwordspacing

\bibitem[Rich and Knight(1991)]{RichKnight91}
E.~Rich and K.~Knight, \emph{Artificial intelligence (2. ed.)}.\hskip 1em plus
  0.5em minus 0.4em\relax McGraw-Hill, 1991.

\bibitem[Fawzi et~al.(2016)Fawzi, Moosavi{-}Dezfooli, and Frossard]{FawziMF16}
\BIBentryALTinterwordspacing
A.~Fawzi, S.~Moosavi{-}Dezfooli, and P.~Frossard, ``Robustness of classifiers:
  from adversarial to random noise,'' in \emph{Advances in Neural Information
  Processing Systems 29: Annual Conference on Neural Information Processing
  Systems 2016, December 5-10, 2016, Barcelona, Spain}, 2016, 2016.
\BIBentrySTDinterwordspacing

\bibitem[Mangasarian(1999)]{Mangasarian99}
\BIBentryALTinterwordspacing
O.~L. Mangasarian, ``Arbitrary-norm separating plane,'' \emph{Oper. Res.
  Lett.}, vol.~24, no. 1-2, pp. 15--23, 1999.
\BIBentrySTDinterwordspacing

\bibitem[Plastria and Carrizosa(2001)]{PlastriaCarrizosa01}
F.~Plastria and E.~Carrizosa, ``Gauge distances and median hyperplanes,''
  \emph{Journal of Optimization Theory and Applications}, vol. 110, no.~1, pp.
  173--182, 2001.

\bibitem[Carlini et~al.(2017)Carlini, Katz, Barrett, and Dill]{CarliniKBD17}
\BIBentryALTinterwordspacing
N.~Carlini, G.~Katz, C.~Barrett, and D.~L. Dill, ``Ground-truth adversarial
  examples,'' \emph{CoRR}, vol. abs/1709.10207, 2017.
\BIBentrySTDinterwordspacing

\bibitem[Peck et~al.(2017)Peck, Roels, Goossens, and Saeys]{PeckRGS17}
\BIBentryALTinterwordspacing
J.~Peck, J.~Roels, B.~Goossens, and Y.~Saeys, ``Lower bounds on the robustness
  to adversarial perturbations,'' in \emph{Advances in Neural Information
  Processing Systems 30: Annual Conference on Neural Information Processing
  Systems 2017, 4-9 December 2017, Long Beach, CA, {USA}}, 2017, 2017.
\BIBentrySTDinterwordspacing

\bibitem[Pohl(1973)]{Pohl73}
\BIBentryALTinterwordspacing
I.~Pohl, ``The avoidance of (relative) catastrophe, heuristic competence,
  genuine dynamic weighting and computational issues in heuristic problem
  solving,'' in \emph{Proceedings of the 3rd International Joint Conference on
  Artificial Intelligence. Standford, CA, USA, August 20-23, 1973}.\hskip 1em
  plus 0.5em minus 0.4em\relax William Kaufmann, 1973, 1973.
\BIBentrySTDinterwordspacing

\bibitem[Pearl and Kim(1982)]{PearlK82}
\BIBentryALTinterwordspacing
J.~Pearl and J.~H. Kim, ``Studies in semi-admissible heuristics,'' \emph{{IEEE}
  Trans. Pattern Anal. Mach. Intell.}, vol.~4, no.~4, pp. 392--399, 1982.
\BIBentrySTDinterwordspacing

\bibitem[Likhachev et~al.(2003)Likhachev, Gordon, and Thrun]{LikhachevGT03}
\BIBentryALTinterwordspacing
M.~Likhachev, G.~J. Gordon, and S.~Thrun, ``{ARA*}: Anytime {A*} with provable
  bounds on sub-optimality,'' in \emph{Advances in Neural Information
  Processing Systems 16 [Neural Information Processing Systems, {NIPS} 2003,
  December 8-13, 2003, Vancouver and Whistler, British Columbia,
  Canada]}.\hskip 1em plus 0.5em minus 0.4em\relax {MIT} Press, 2003, 2003.
\BIBentrySTDinterwordspacing

\bibitem[Pedregosa et~al.(2011)Pedregosa, Varoquaux, Gramfort, Michel, Thirion,
  Grisel, Blondel, Prettenhofer, Weiss, Dubourg, Vanderplas, Passos,
  Cournapeau, Brucher, Perrot, and Duchesnay]{scikit-learn}
F.~Pedregosa, G.~Varoquaux, A.~Gramfort, V.~Michel, B.~Thirion, O.~Grisel,
  M.~Blondel, P.~Prettenhofer, R.~Weiss, V.~Dubourg, J.~Vanderplas, A.~Passos,
  D.~Cournapeau, M.~Brucher, M.~Perrot, and E.~Duchesnay, ``Scikit-learn:
  Machine learning in {P}ython,'' \emph{Journal of Machine Learning Research},
  vol.~12, pp. 2825--2830, 2011.

\bibitem[Kluyver et~al.(2016)Kluyver, Ragan-Kelley, P{\'e}rez, Granger,
  Bussonnier, Frederic, Kelley, Hamrick, Grout, Corlay, Ivanov, Avila, Abdalla,
  and Willing]{jupyter}
T.~Kluyver, B.~Ragan-Kelley, F.~P{\'e}rez, B.~Granger, M.~Bussonnier,
  J.~Frederic, K.~Kelley, J.~Hamrick, J.~Grout, S.~Corlay, P.~Ivanov, D.~Avila,
  S.~Abdalla, and C.~Willing, ``Jupyter notebooks -- a publishing format for
  reproducible computational workflows,'' in \emph{Positioning and Power in
  Academic Publishing: Players, Agents and Agendas}.\hskip 1em plus 0.5em minus
  0.4em\relax IOS Press, 2016, 2016.

\bibitem[Tange(2011)]{gnu-parallel}
\BIBentryALTinterwordspacing
O.~Tange, ``Gnu parallel - the command-line power tool,'' \emph{;login: The
  USENIX Magazine}, vol.~36, no.~1, pp. 42--47, Feb 2011.
\BIBentrySTDinterwordspacing

\bibitem[Gilani et~al.(2017)Gilani, Kochmar, and Crowcroft]{GilaniKC17}
\BIBentryALTinterwordspacing
Z.~Gilani, E.~Kochmar, and J.~Crowcroft, ``Classification of twitter accounts
  into automated agents and human users,'' in \emph{Proceedings of the 2017
  IEEE/ACM International Conference on Advances in Social Networks Analysis and
  Mining 2017}, ser. ASONAM '17.\hskip 1em plus 0.5em minus 0.4em\relax New
  York, NY, USA: ACM, 2017, 2017.
\BIBentrySTDinterwordspacing

\bibitem[Wang et~al.(2014)Wang, Cai, Nithyanand, Johnson, and
  Goldberg]{WangCNJG14}
\BIBentryALTinterwordspacing
T.~Wang, X.~Cai, R.~Nithyanand, R.~Johnson, and I.~Goldberg, ``Effective
  attacks and provable defenses for website fingerprinting,'' in
  \emph{Proceedings of the 23rd {USENIX} Security Symposium, San Diego, CA,
  USA, August 20-22, 2014.}\hskip 1em plus 0.5em minus 0.4em\relax {USENIX}
  Association, 2014, 2014.
\BIBentrySTDinterwordspacing

\bibitem[Hayes and Danezis(2016)]{HayesD16}
\BIBentryALTinterwordspacing
J.~Hayes and G.~Danezis, ``k-fingerprinting: {A} robust scalable website
  fingerprinting technique,'' in \emph{25th {USENIX} Security Symposium,
  {USENIX} Security 16, Austin, TX, USA, August 10-12, 2016.}\hskip 1em plus
  0.5em minus 0.4em\relax {USENIX} Association, 2016, 2016.
\BIBentrySTDinterwordspacing

\bibitem[Sirinam et~al.(2018)Sirinam, Imani, Ju{\'{a}}rez, and
  Wright]{SirinamIJW18}
\BIBentryALTinterwordspacing
P.~Sirinam, M.~Imani, M.~Ju{\'{a}}rez, and M.~Wright, ``Deep fingerprinting:
  Undermining website fingerprinting defenses with deep learning,'' in
  \emph{Proceedings of the 2018 {ACM} {SIGSAC} Conference on Computer and
  Communications Security, {CCS} 2018, Toronto, ON, Canada, October 15-19,
  2018}.\hskip 1em plus 0.5em minus 0.4em\relax {ACM}, 2018, 2018.
\BIBentrySTDinterwordspacing

\bibitem[Rimmer et~al.(2018)Rimmer, Preuveneers, Ju{\'{a}}rez, van Goethem, and
  Joosen]{RimmerPJGJ18}
\BIBentryALTinterwordspacing
V.~Rimmer, D.~Preuveneers, M.~Ju{\'{a}}rez, T.~van Goethem, and W.~Joosen,
  ``Automated website fingerprinting through deep learning,'' in \emph{25th
  Annual Network and Distributed System Security Symposium, {NDSS} 2018, San
  Diego, California, USA, February 18-21, 2018}.\hskip 1em plus 0.5em minus
  0.4em\relax The Internet Society, 2018, 2018.
\BIBentrySTDinterwordspacing

\bibitem[Dingledine et~al.(2004)Dingledine, Mathewson, and
  Syverson]{DingledineMS04}
\BIBentryALTinterwordspacing
R.~Dingledine, N.~Mathewson, and P.~F. Syverson, ``Tor: The second-generation
  onion router,'' in \emph{Proceedings of the 13th {USENIX} Security Symposium,
  August 9-13, 2004, San Diego, CA, {USA}}.\hskip 1em plus 0.5em minus
  0.4em\relax {USENIX}, 2004, 2004.
\BIBentrySTDinterwordspacing

\bibitem[Rescorla et~al.(2018)Rescorla, Oku, Sullivan, and Wood]{ietf-tls-esni}
\BIBentryALTinterwordspacing
E.~Rescorla, K.~Oku, N.~Sullivan, and C.~Wood, ``{Encrypted Server Name
  Indication for TLS 1.3},'' Working Draft, IETF Secretariat, Internet-Draft
  draft-ietf-tls-esni-02, October 2018,
  \url{http://www.ietf.org/internet-drafts/draft-ietf-tls-esni-02.txt}.
\BIBentrySTDinterwordspacing

\bibitem[Rescorla(2018)]{RFC8446}
E.~Rescorla, ``{The Transport Layer Security (TLS) Protocol Version 1.3},''
  Internet Requests for Comments, RFC Editor, RFC 8446, August 2018.

\bibitem[Cherubin(2017)]{Cherubin17}
\BIBentryALTinterwordspacing
G.~Cherubin, ``Bayes, not na{\"{\i}}ve: Security bounds on website
  fingerprinting defenses,'' \emph{PoPETs}, vol. 2017, no.~4, pp. 215--231,
  2017.
\BIBentrySTDinterwordspacing

\bibitem[Dalvi et~al.(2004)Dalvi, Domingos, Mausam, Sanghai, and
  Verma]{DalviDMSV04}
\BIBentryALTinterwordspacing
N.~N. Dalvi, P.~M. Domingos, Mausam, S.~K. Sanghai, and D.~Verma, ``Adversarial
  classification,'' in \emph{Proceedings of the Tenth {ACM} {SIGKDD}
  International Conference on Knowledge Discovery and Data Mining, Seattle,
  Washington, USA, August 22-25, 2004}.\hskip 1em plus 0.5em minus 0.4em\relax
  {ACM}, 2004, 2004.
\BIBentrySTDinterwordspacing

\bibitem[Hosseini et~al.(2017)Hosseini, Kannan, Zhang, and
  Poovendran]{HosseiniKZP17}
\BIBentryALTinterwordspacing
H.~Hosseini, S.~Kannan, B.~Zhang, and R.~Poovendran, ``Deceiving google's
  perspective {API} built for detecting toxic comments,'' \emph{CoRR}, vol.
  abs/1702.08138, 2017.
\BIBentrySTDinterwordspacing

\bibitem[Alzantot et~al.(2018)Alzantot, Sharma, Elgohary, Ho, Srivastava, and
  Chang]{AlzantotSEHSC18}
\BIBentryALTinterwordspacing
M.~Alzantot, Y.~Sharma, A.~Elgohary, B.~Ho, M.~B. Srivastava, and K.~Chang,
  ``Generating natural language adversarial examples,'' in \emph{Proceedings of
  the 2018 Conference on Empirical Methods in Natural Language Processing,
  Brussels, Belgium, October 31 - November 4, 2018}.\hskip 1em plus 0.5em minus
  0.4em\relax Association for Computational Linguistics, 2018, 2018.
\BIBentrySTDinterwordspacing

\bibitem[Dang et~al.(2017)Dang, Huang, and Chang]{DangHC17}
\BIBentryALTinterwordspacing
H.~Dang, Y.~Huang, and E.~Chang, ``Evading classifiers by morphing in the
  dark,'' in \emph{Proceedings of the 2017 {ACM} {SIGSAC} Conference on
  Computer and Communications Security, {CCS} 2017, Dallas, TX, USA, October 30
  - November 03, 2017}.\hskip 1em plus 0.5em minus 0.4em\relax {ACM}, 2017,
  2017.
\BIBentrySTDinterwordspacing

\bibitem[Xu et~al.(2016)Xu, Qi, and Evans]{XuQE16}
\BIBentryALTinterwordspacing
W.~Xu, Y.~Qi, and D.~Evans, ``Automatically evading classifiers: {A} case study
  on {PDF} malware classifiers,'' in \emph{23rd Annual Network and Distributed
  System Security Symposium, {NDSS} 2016, San Diego, California, USA, February
  21-24, 2016}.\hskip 1em plus 0.5em minus 0.4em\relax The Internet Society,
  2016, 2016.
\BIBentrySTDinterwordspacing

\bibitem[Wu et~al.(2018)Wu, Wicker, Ruan, Huang, and Kwiatkowska]{WuWRHK18}
\BIBentryALTinterwordspacing
M.~Wu, M.~Wicker, W.~Ruan, X.~Huang, and M.~Kwiatkowska, ``A game-based
  approximate verification of deep neural networks with provable guarantees,''
  \emph{CoRR}, vol. abs/1807.03571, 2018.
\BIBentrySTDinterwordspacing

\bibitem[Kulynych(2017)]{textfool}
B.~Kulynych, ``textfool: Plausible looking adversarial examples for text
  classification,'' Jul 2017.

\bibitem[Lee et~al.(2016)Lee, Sugiyama, von Luxburg, Guyon, and
  Garnett]{DBLP:conf/nips/2016}
\BIBentryALTinterwordspacing
D.~D. Lee, M.~Sugiyama, U.~von Luxburg, I.~Guyon, and R.~Garnett, Eds.,
  \emph{Advances in Neural Information Processing Systems 29: Annual Conference
  on Neural Information Processing Systems 2016, December 5-10, 2016,
  Barcelona, Spain}, 2016.
\BIBentrySTDinterwordspacing

\bibitem[Guyon et~al.(2017)Guyon, von Luxburg, Bengio, Wallach, Fergus,
  Vishwanathan, and Garnett]{DBLP:conf/nips/2017}
I.~Guyon, U.~von Luxburg, S.~Bengio, H.~M. Wallach, R.~Fergus, S.~V.~N.
  Vishwanathan, and R.~Garnett, Eds., \emph{Advances in Neural Information
  Processing Systems 30: Annual Conference on Neural Information Processing
  Systems 2017, 4-9 December 2017, Long Beach, CA, {USA}}, 2017.

\end{thebibliography}
}

\begin{appendices}
\section{Proof of \Stmtref{thm:admissibility}}
\label{app:admissibility-proof}
Observe that if $F(\vx) = t$, the heuristic $\dist(\vx)
= 0$, and hence is trivially admissible.  Indeed, it cannot overestimate $\graphcost(\vx, \optim{\vx})$
due to the fact that $\edgecost(a, b) \geq 0$ and $\graphcost(a, b) \geq 0$ for any $a, b \in V$.

It is therefore sufficient to show that if $F(\vx) \neq t$, the lower bound on adversarial
robustness at $x$ over $\domainclosure$ never overestimates $\graphcost(\vx, \optim{\vx})$:
\begin{equation}\label{eq:lowbound-less-than-graphcost}
    \distlowbound(\vx') \leq \graphcost(\vx, \optim{\vx})
\end{equation}

The following sequence holds:
\[
    \begin{aligned}
        \dist(\vx) & \leq \norm{\vx - \optim{\vx}} \\
        & \leq \graphcost(\vx, \optim{\vx}) \\
    \end{aligned}
\]

The first inequality is by definition of $\dist(\vx)$ (see \Eqref{eq:robustness}).  Indeed, since
$\dist(\vx)$ is a norm of the smallest adversarial perturbation $\Delta$ over $\domainclosure$,
$\Delta$ is smaller than the distance from $\vx$ to any other $\vx' \in \sX \subseteq \domainclosure$
that also flips the decision of the target classifier:
\[
    \dist(\vx) = \norm{\Delta} \leq \norm{\vx - \vx'} \quad \text{\footnotesize (for \emph{any} $\vx' \in \sX$ s.t.
    $F(\vx') = t$)}
\]

By \Eqref{eq:graph-cost}, $\graphcost(\vx, \optim{\vx})$ is a path cost for some path:
\[
    \begin{aligned}
        \graphcost(\vx, \optim{\vx}) & = \pathcost(\vx \rightarrow \vv_1 \rightarrow \ldots
\rightarrow \vv_{n-1} \rightarrow \optim{\vx}) \\
        & = \norm{\vx - \vv_1} + \sum_{i=1}^{n-2} \norm{\vv_i - \vv_{i+1}} + \norm{\vv_{n-1} -
        \optim{\vx}}
    \end{aligned}
\]

By triangle property of the norm, the second inequality holds:
\[
    \begin{aligned}
        \norm{\vx - \optim{\vx}} & \leq \norm{\vx - \vv_1} + \sum_{i=1}^{n-2} \norm{\vv_i - \vv_{i+1}} +
        \norm{\vv_{n-1} - \optim{\vx}} \\
            & = \graphcost(x, \optim{\vx})
    \end{aligned}
\]

Hence, $\dist(\vx) \leq \graphcost(\vx, \optim{\vx})$, which implies
\Eqref{eq:lowbound-less-than-graphcost}, and concludes the proof.

\section{Derivation of the heuristic approximation for non-linear models}
\label{sec:non-linear-heuristic-deriv}
W.l.o.g, assume that the decision threshold of the target classifier is $\scorethresh = 0$. For an initial $\vx
\in \sR^m$, the smallest adversarial perturbation $\Delta \in \sR^m$ puts $\vx + \Delta$ \emph{on}
the decision boundary: $f(\vx + \Delta) = \scorethresh = 0$.

Let $\tilde f(\vx + \Delta)$ be the first-order Taylor approximation of $f$ at $\vx + \Delta$:
\[\tilde f(\vx + \Delta) = f(\vx) + \nabla_{\vx} f(\vx) \cdot \Delta\]

We want to estimate $\Delta$ assuming $\tilde f(\vx + \Delta) = f(\vx + \Delta) = 0$. By H\"older's inequality,
\[|f(\vx)| = |\nabla_{\vx} f(\vx) \cdot \Delta| \leq \dualnorm{\nabla_{\vx} f(\vx)} \norm{\Delta} \]

Hence, assuming that $\tilde f(\vx + \Delta) = 0$, the $p$-norm of the smallest perturbation has the
following lower bound:
\[\norm{\Delta} \geq \frac{|f(\vx)|}{\dualnorm{\nabla_{\vx} f(\vx)}} \]

We can use the right-hand side as an approximation of the lower bound on $\dist(\vx)$.

Note that for a linear model $f(\vx) = \vw \cdot \vx + b$ the first-order approximation $\tilde
f(\vx + \Delta)$ is exact. Hence, the bound implies \Eqref{eq:dist-decision-boundary}:
\[\norm{\Delta} \geq \frac{|f(\vx)|}{\dualnorm{\vw}}\]

\section{Supplementary figures}
\label{app:figures}
The rest of the document contains supplementary figures.

\begin{sidewaystable*}
    \caption{Instantiations of existing attacks within the graphical framework}%
\label{tab:comparison}
\begin{center}
\resizebox{\columnwidth}{!}{
    \begin{tabular}{lcclcllc}
    \toprule
    \textbf{Attack} & \textbf{Domain} & \textbf{Adversary knowledge} & \textbf{Expansions} & \textbf{Cost} & \textbf{Heuristic} & \textbf{Search algorithm} & \textbf{Admissibility} \\

    \midrule

    \multirow{2}{*}{\citet{PapernotMSH16}}
    & \multirow{2}{*}{Text}
    & \multirow{2}{*}{White-box}
    & \multirow{2}{*}{\shortstack[l]{Word substitutions}}
    & \multirow{2}{*}{Number of substitutions}
    & \multirow{2}{*}{Forward gradient-based}
    & \multirow{2}{*}{Hill climbing}
    & \multirow{2}{*}{\nope} \\

    &
    &
    &
    &
    &
    &
    &  \\
        [0.7cm]

    \multirow{3}{*}{\citet{textfool}}
    & \multirow{3}{*}{Text}
    & \multirow{3}{*}{White-box}
    & \multirow{3}{*}{\shortstack[l]{Word substitutions, \\ character substitutions,\\ insertions, deletions}}
    & \multirow{3}{*}{Semantic dissimilarity }
    & \multirow{3}{*}{Forward gradient-based}
    & \multirow{3}{*}{Best-first}
    & \multirow{3}{*}{\nope} \\

    &
    &
    &
    &
    &
    &
    &  \\

    &
    &
    &
    &
    &
    &
    &  \\
    [0.7cm]

    \multirow{3}{*}{\citet{LiangSBLS18}}
        & \multirow{3}{*}{Text}
        & \multirow{3}{*}{White-box}
        & \multirow{3}{*}{\shortstack[l]{Insertion of common phrases,\\ word removal, character substitution,\\homoglyph insertions}}

        & \multirow{3}{*}{---}
        & \multirow{3}{*}{Forward gradient-based}
    & \multirow{3}{*}{Hill climbing}
    & \multirow{3}{*}{\nope} \\

    &
    &
    &
    &
    &
    &
    &  \\

    &
    &
    &
    &
    &
    &
    &  \\
    [0.7cm]

    \multirow{2}{*}{\citet{EbrahimiRLD18}}
        & \multirow{2}{*}{Text}
        & \multirow{2}{*}{White-box}
    & \multirow{2}{*}{\shortstack[l]{Character/word substitution,\\ insertion, deletion}}
        & \multirow{2}{*}{---}
        & \multirow{2}{*}{Forward gradient-based}
        & \multirow{2}{*}{Best-first beam search}
    & \multirow{2}{*}{\nope} \\

    &
    &
    &
    &
    &
    &
    &  \\
    [0.7cm]

    \citet{GaoLSQ18}
    & Text
    & Black-box
    & Character substitution, insertion, deletion
    & ---
    & Confidence-based
    & Hill climbing
    & \nope \\
    [0.3cm]

    \midrule

    \multirow{2}{*}{\citet{GrossePM0M16}}
        & \multirow{2}{*}{Malware}
        & \multirow{2}{*}{White-box}
        & \multirow{2}{*}{Bit flips}
        & \multirow{2}{*}{$\lp$}
        & \multirow{2}{*}{Forward gradient-based}
    & \multirow{2}{*}{Hill climbing}
    & \multirow{2}{*}{\nope} \\
    &
    &
    &
    &
    &
    &
    &  \\

        \midrule

        \multirow{2}{*}{\citet{JiaG18}}
        & \multirow{2}{*}{{\footnotesize App recom\-men\-dations}}
            & \multirow{2}{*}{White-box}
            & \multirow{2}{*}{Item addition, modification}
            & \multirow{2}{*}{$\lp$}
            & \multirow{2}{*}{Forward gradient-based}
        & \multirow{2}{*}{Hill climbing}
        & \multirow{2}{*}{\nope} \\
        &
    &
    &
    &
    &
    &
    &  \\

        \midrule
        \multirow{2}{*}{\citet{OverdorfKBTG18}}
            & \multirow{2}{*}{Credit scoring}
            & \multirow{2}{*}{White-box}
            & \multirow{2}{*}{Category modifications}
            & \multirow{2}{*}{Number of modifications}
            & \multirow{2}{*}{---}
        & \multirow{2}{*}{Uniform-cost}
        & \multirow{2}{*}{\yup} \\

        &
    &
    &
    &
    &
    &
    &  \\

        \midrule
        \multirow{2}{*}{\textbf{Ours}}
            & \multirow{2}{*}{*}
            & \multirow{2}{*}{White-box}
            & \multirow{2}{*}{*}
            & \multirow{2}{*}{Graph path cost with $\norm{\cdot}$ edge costs}
        & \multirow{2}{*}{\shortstack[l]{Lower bound on minimal \\adversarial cost over continuous domain}}
        & \multirow{2}{*}{\shortstack[l]{\astar}}
        & \multirow{2}{*}{\yup} \\

    &
    &
    &
    &
    &
    &
    &  \\

    \bottomrule
    \end{tabular}}
\end{center}

\end{sidewaystable*}

\begin{sidewaystable*}
    \caption{Examples of feature transformations that cause the classifier (<1000 followers,
bucketization parameter 20) to misclassify a Twitter bot account as a human account. The first line
in every pair corresponds to intial feature values, the second line shows the feature values in
a MAC adversarial example. Transformed feature values are emphasized in bold. Confidence is given
for the `human' label}%
\label{tab:bots-examples}
\centering
\resizebox{\textwidth}{!}{
    \begin{tabular}{lllllllll}
        \toprule
        Confidence  & \texttt{source\_identity}         & \texttt{user\_tweeted} & \texttt{user\_retweeted} & \texttt{user\_favourited} & \texttt{user\_replied} & \texttt{lists\_per\_user}   & \texttt{age\_of\_account\_in\_days} & \texttt{urls\_count}   \\
        \midrule
        0.120752 &  \textbf{[other, browser, mobile]} & (6.0, 7.0]             & [0, 1.0]                 & \textbf{(52.0, 207.8]}    & [0, 1.0]               & (0.011, 0.0143]             & (2127.598, 2264.455]                & (5.0, 6.0]             \\
        0.544546 &\textbf{[browser, mobile]}        & (6.0, 7.0]             & [0, 1.0]                 & \textbf{(207.8, 476.143]} & [0, 1.0]               & (0.011, 0.0143]             & (2127.598, 2264.455]                & (5.0, 6.0]             \\
        \midrule
        0.008606 & \textbf{[other]}                  & [0, 1.0]               & \textbf{(19.0, 32.0]}    & [0, 4.0]                  & (26.0, 42.0]           & \textbf{(0.024, 0.0329]}    & (766.122, 849.108]                  & (16.0, 24.2]           \\
        0.526738 & \textbf{[mobile, osn]}            & [0, 1.0]               & \textbf{(12.0, 19.0]}    & [0, 4.0]                  & (26.0, 42.0]           & \textbf{(0.0188, 0.024]}    & (766.122, 849.108]                  & (16.0, 24.2]           \\
        \midrule
        0.118470 & \textbf{[other, browser, mobile]} & (20.0, 27.0]           & \textbf{(19.0, 32.0]}    & (35990.095, 211890.704]   & (18.0, 26.0]           & (0.024, 0.0329]             & (2450.073, 3332.802]                & (16.0, 24.2]           \\
        0.578171 & \textbf{[browser, mobile]}        & (20.0, 27.0]           & \textbf{(12.0, 19.0]}    & (35990.095, 211890.704]   & (18.0, 26.0]           & (0.024, 0.0329]             & (2450.073, 3332.802]                & (16.0, 24.2]           \\
        \midrule
        0.050031 & \textbf{[marketing]}              & \textbf{(55.6, 766.0]} & [0, 1.0]                 & (772.24, 1164.881]        & [0, 1.0]               & (0.05, 0.0962]              & (1384.307, 1492.455]                & (56.0, 806.0]          \\
        0.511897 & \textbf{[mobile, osn]}            & \textbf{(27.0, 55.6]}  & [0, 1.0]                 & (772.24, 1164.881]        & [0, 1.0]               & (0.05, 0.0962]              & (1384.307, 1492.455]                & (56.0, 806.0]          \\
        \midrule
        0.009900 & \textbf{[automation]}             & (13.0, 16.0]           & [0, 1.0]                 & (4.0, 52.0]               & \textbf{[0, 1.0]}      & (0.05, 0.0962]              & \textbf{(1177.447, 1269.787]}       & (12.0, 16.0]           \\
        0.512774 & \textbf{[mobile, osn]}            & (13.0, 16.0]           & [0, 1.0]                 & (4.0, 52.0]               & \textbf{(1.0, 2.0]}    & (0.05, 0.0962]              & \textbf{(1269.787, 1384.307]}       & (12.0, 16.0]           \\
        \midrule
        0.012389 & \textbf{[automation]}             & (16.0, 20.0]           & [0, 1.0]                 & [0, 4.0]                  & \textbf{[0, 1.0]}      & \textbf{(0.00373, 0.00485]} & (503.551, 645.233]                  & (16.0, 24.2]           \\
        0.012389 & \textbf{[mobile]}                 & (16.0, 20.0]           & [0, 1.0]                 & [0, 4.0]                  & \textbf{(1.0, 2.0]}    & \textbf{(0.00485, 0.00594]} & (503.551, 645.233]                  & (16.0, 24.2]           \\
        \midrule
        0.003755 & \textbf{[automation]}             & \textbf{(55.6, 766.0]} & [0, 1.0]                 & [0, 4.0]                  & \textbf{[0, 1.0]}      & \textbf{(0.024, 0.0329]}    & (1939.803, 2127.598]                & (56.0, 806.0]          \\
        0.530595 & \textbf{[mobile]}                 & \textbf{(27.0, 55.6]}  & [0, 1.0]                 & [0, 4.0]                  & \textbf{(1.0, 2.0]}    & \textbf{(0.0143, 0.0188]}   & (1939.803, 2127.598]                & (56.0, 806.0]          \\
        \midrule
        0.024658 & \textbf{[marketing]}              & \textbf{(55.6, 766.0]} & (2.0, 3.0]               & [0, 4.0]                  & \textbf{[0, 1.0]}      & (0.0962, 62.702]            & (849.108, 956.554]                  & \textbf{(56.0, 806.0]} \\
        0.543155 & \textbf{[mobile]}                 & \textbf{(27.0, 55.6]}  & (2.0, 3.0]               & [0, 4.0]                  & \textbf{(1.0, 2.0]}    & (0.0962, 62.702]            & (849.108, 956.554]                  & \textbf{(24.2, 56.0]}  \\
        \midrule
        0.004241 & \textbf{[other]}                  & \textbf{(55.6, 766.0]} & [0, 1.0]                 & (4.0, 52.0]               & \textbf{[0, 1.0]}      & (0.05, 0.0962]              & \textbf{(1177.447, 1269.787]}       & \textbf{(56.0, 806.0]} \\
        0.501168 & \textbf{[mobile, osn]}            & \textbf{(27.0, 55.6]}  & [0, 1.0]                 & (4.0, 52.0]               & \textbf{(1.0, 2.0]}    & (0.05, 0.0962]              & \textbf{(1269.787, 1384.307]}       & \textbf{(24.2, 56.0]}  \\
        \midrule
        0.017455 & \textbf{[other]}                  & (7.0, 9.0]             & [0, 1.0]                 & [0, 4.0]                  & \textbf{[0, 1.0]}      & (0.05, 0.0962]              & (17.321, 209.213]                   & (7.0, 9.0]             \\
        0.509626 & \textbf{[mobile, osn]}            & (7.0, 9.0]             & [0, 1.0]                 & [0, 4.0]                  & \textbf{(1.0, 2.0]}    & (0.05, 0.0962]              & (17.321, 209.213]                   & (7.0, 9.0]             \\
        \bottomrule
        \multicolumn{9}{p{\textwidth}}{
        \vspace{.25em}
        \footnotesize
        Features that were not changed in any adversarial example are not shown.
        }
    \end{tabular}
}

\end{sidewaystable*}

\begin{figure*}[t]
    \centering
        \includegraphics[width=0.99\textwidth]{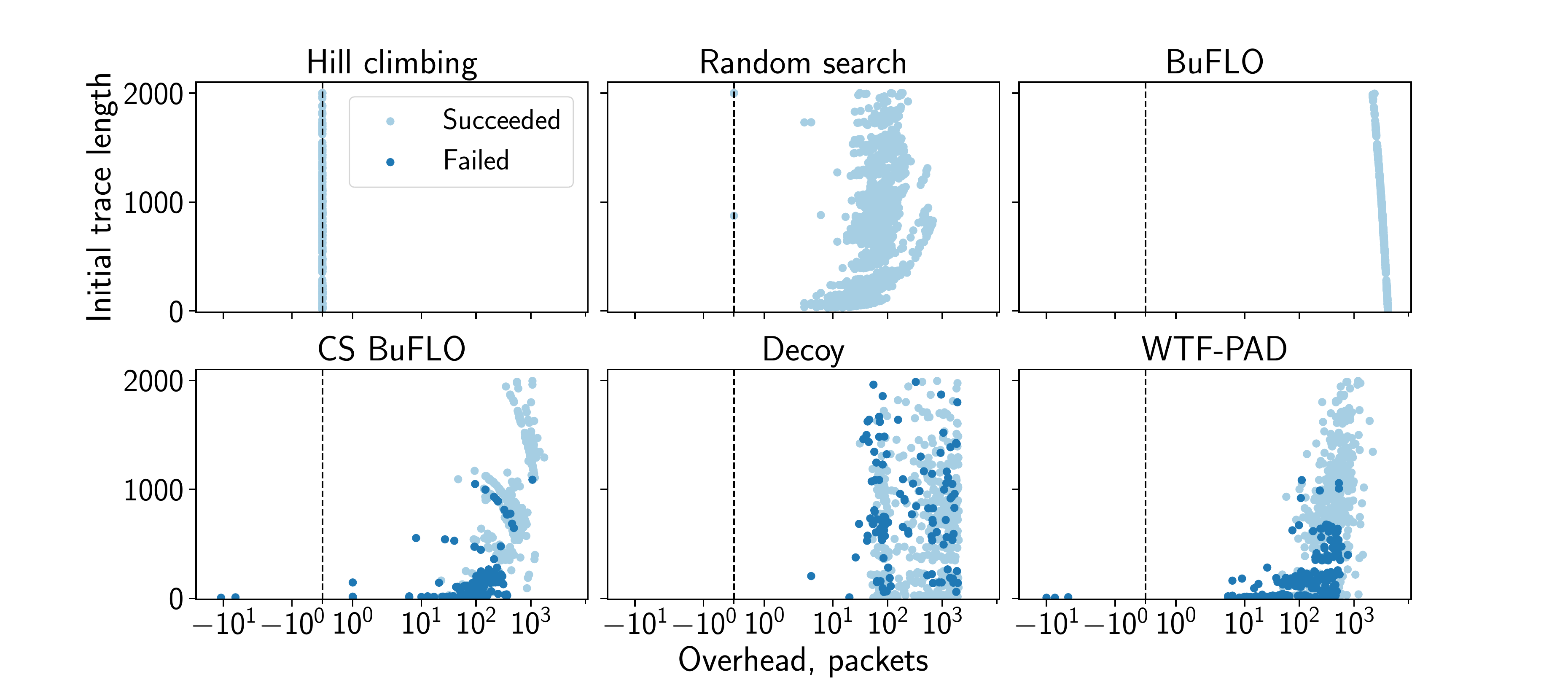}
        \caption{Overhead of WF defenses compared to adversarial examples found with hill-climbing search (x-axis
        is bi-symmetrically logarithmic)}
    \label{fig:wfp-defenses-delta}
\end{figure*}

\end{appendices}

\end{document}